\documentclass[10pt,twocolumn,letterpaper]{article}

\usepackage{iccv}
\usepackage{times}
\usepackage{epsfig}
\usepackage{graphicx}
\usepackage{amsmath}
\usepackage{amssymb}
\usepackage{times}
\usepackage{epsfig}
\usepackage{graphicx}
\usepackage{amsmath,amsthm,amssymb}
\usepackage{engord}
\usepackage{bm}
\usepackage{amstext}
\usepackage{graphicx}
\usepackage{comment}
\usepackage{bbm}
\usepackage{color}
\usepackage[table]{xcolor}
\usepackage{ctable}
\usepackage{booktabs}
\usepackage{multirow}
\usepackage{mathtools}
\usepackage{setspace}
\usepackage{pifont}
\usepackage{amsmath}
\usepackage{amssymb}
\usepackage{dsfont}
\usepackage[inline]{enumitem}

\definecolor{Gray}{gray}{0.9}
\definecolor{White}{gray}{1}

\newcommand\vC{\textit{VC}}
\newcommand\cR{\textit{CR}}
\newcommand\aE{\textit{AE}}

\usepackage[ruled,vlined,linesnumbered]{algorithm2e}

\usepackage[pagebackref=true,breaklinks=true,letterpaper=true,colorlinks,bookmarks=false]{hyperref}

\iccvfinalcopy 

\def\iccvPaperID{2919} 

\begin{document}

\title{CPF: Learning a Contact Potential Field to Model the Hand-Object Interaction}

\author{{Lixin Yang},~ {Xinyu Zhan},~ {Kailin Li},~ {Wenqiang Xu},~ {Jiefeng Li},~ {Cewu Lu}\\
{Shanghai Jiao Tong University, Shanghai, China}\\
{\tt\small \{{siriusyang}, {kelvin34501}, {kailinli}, {vinjohn},  {ljf\_likit}, {lucewu}\}@{sjtu.edu.cn}}
}

\maketitle

\begin{abstract}
   Modeling the hand-object (HO) interaction not only requires estimation of the HO pose, but also pays attention to the contact due to their interaction.
   Significant progress has been made in estimating hand and object separately with deep learning methods, simultaneous HO pose estimation and contact modeling has not yet been fully explored.
   In this paper, we present an explicit contact representation namely Contact Potential Field (CPF), and a learning-fitting hybrid framework namely MIHO to Modeling the Interaction of Hand and Object. In CPF, we treat each contacting HO vertex pair as a spring-mass system. Hence the whole system forms a potential field with minimal elastic energy at the grasp position. Extensive experiments on the two commonly used benchmarks have demonstrated that our method can achieve state-of-the-art in several reconstruction metrics, and allow us to produce more physically plausible HO pose even when the ground-truth exhibits severe interpenetration or disjointedness. Our code is available at \url{https://github.com/lixiny/CPF}.
\end{abstract}


\vspace{-1em}\section{Introduction}
It is essential to model hand-object interaction from a single image for understanding the human activities, in which simulating a physically plausible grasp is also crucial for VR$/$AR, teleoperation, and grasping applications.
Given an image as input, the problem aims to not only estimate proper hand-object pose but also to recover a natural grasp configuration.
While estimating hand \cite{moon2020deephandmesh, kulon2020weakly, zhou2020monocular, boukhayma20193d, ge20193d, yang2020bihand} or object \cite{groueix2018papier, hasson2020leveraging,  fan2017point,
wang2018pixel2mesh, wu2017marrnet} alone has made a considerable success over the past decades,
simultaneously estimating hand-object pose \cite{hasson2019learning, tekin2019h+, hasson2020leveraging, kokic2019learning, doosti2020hope} with interaction has only emerged in the past few years.

Previous works on joint hand-object estimation usually treat the contact as a result of the correct pose estimation \cite{hasson2020leveraging, kokic2020learning, rogez2015understanding}.
Apparently, if the hand and object can be perfectly recovered, the contact between them will also be satisfied. Yet, such perfection cannot be achieved in practice.
Since contact can provide rich cues to guide accurate pose and natural grasp, more attention has been recently drawn to the contact modeling \cite{brahmbhatt2019contactdb, brahmbhatt2020contactpose} and contact representation \cite{karunratanakul2020grasping, brahmbhatt2019contactgrasp}.
And several contact datasets \cite{brahmbhatt2019contactdb, brahmbhatt2020contactpose, GRAB:2020} have been released to the community.
However, a solution of properly integrating contact modeling into the current hand-object pose estimation pipeline has remained an open research question.
The existing methods either exploit distance-based attraction and repulsion \cite{hasson2019learning, karunratanakul2020grasping} to mitigate disjointedness and interpenetration, or refine the predicted pose in virtue of physics simulators \cite{kokic2019learning, kokic2020learning, garcia2020physics}. While the both solutions are considered to be irrelevant to contact \textit{semantics}, which we will explain later, the latter solutions also lack flexibility on hand pose and shape.


\begin{figure}[t]
   \begin{center}
      \includegraphics[width=1.0\linewidth]{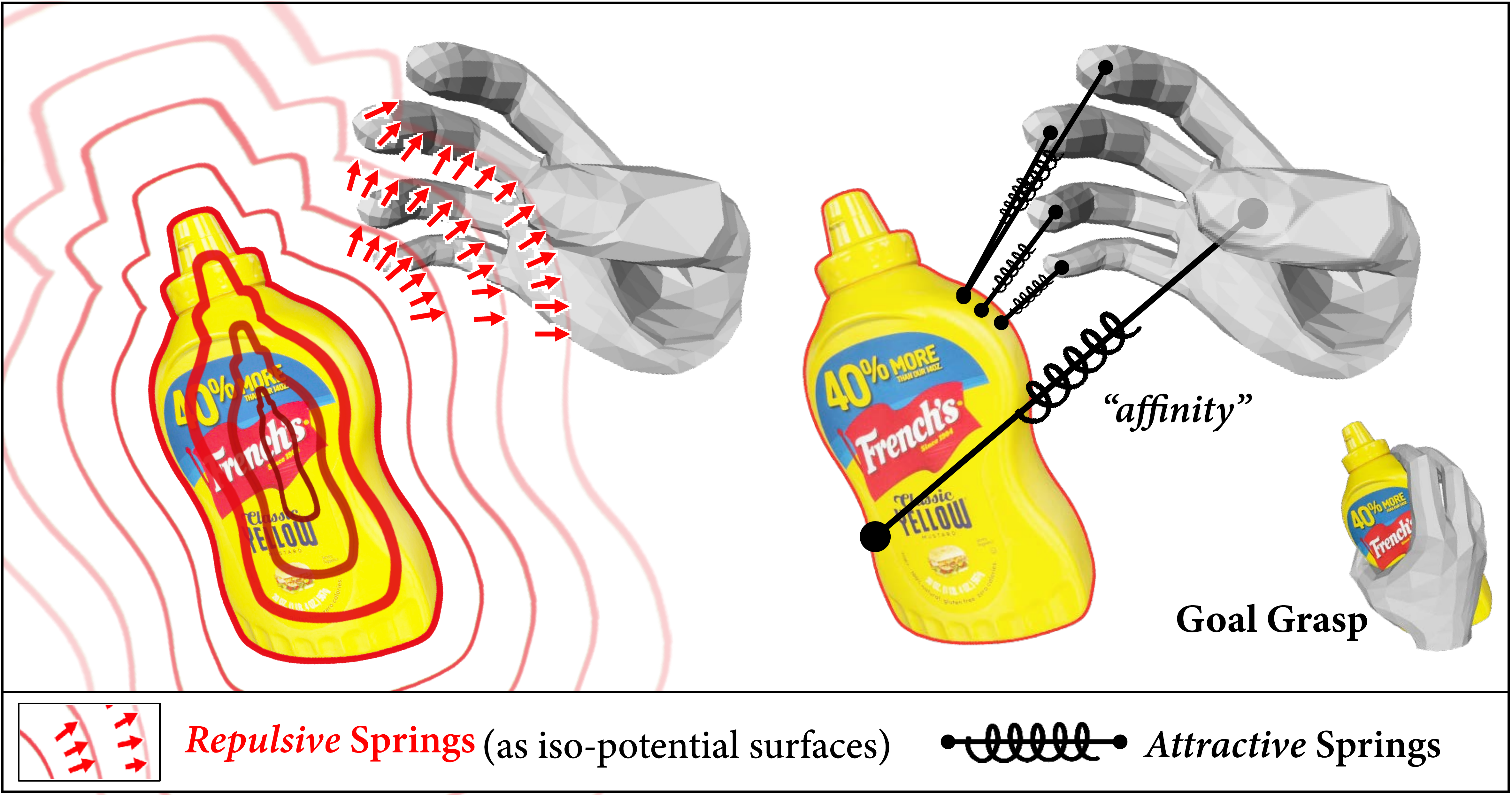}
   \end{center}
      \vspace{-0.5em}\caption{\textbf{Illustration of the proposed Contact Potential Field.} The contacts between hand and object vertices are modeled as the attractive (right) and repulsive (left) springs that connect paired vertex on them.} \vspace{-1.0em}
   \label{fig:capture}
   \label{fig:cpf}
\end{figure}

To model the contact, we propose an explicit representation named \textbf{C}ontact \textbf{P}otential \textbf{F}ield (\textbf{CPF}, \S\ref{sec:cpf}). It is built upon the idea that the contact between a hand and an object mesh under grasp configuration is multi-point contact, which involves multiple hand-object vertex pair affinities. These affinities are regarded as the contact \textit{semantics}, which depict the pairing of the hand-object vertices that come into contact with each other during the interaction.
When noisy predicted hand and object are disjointed from each other, we shall apply an attraction to pull these vertex pairs close; While the hand and object are intersected, we shall have a repulsion to push them away.
Contacts of those affinitive vertex pairs are the result of equilibrium between the attraction and repulsion.
In this paper, we treat each contacting HO vertex pair as a spring-mass system.
First, the two end-points of spring is a counterpart of the two HO vertices in affinity.
Second, the spring's elastic property is another counterpart of the intensity of the vertex pair affinity.
In this way, we can model the HO interaction with a potential field, as we call it CPF,
which is determined by minimal elastic energy at the grasp position.
Therefore, estimating the HO pose under contact is equivalent to minimizing the elastic energy inside CPF.
Representing contact as CPF has two main advantages.
First, compared with contact heuristic with proximity metrics \cite{antotsiou2018task, tzionas2016capturing} or distance field \cite{karunratanakul2020grasping, brahmbhatt2019contactgrasp},
CPF is able to assign per-vertex contact \textit{semantics} (contact points on different hand part) to object mesh.
Second, by minimizing the elastic energy, CPF can uniformly avoid interpenetration and control the disjointedness. Based on CPF, we also propose a novel learning-fitting hybrid framework namely  for \textbf{M}odeling the \textbf{I}nteraction of \textbf{H}and and \textbf{O}bject, as we call it \textbf{MIHO} (\S\ref{sec:approach}).

\begin{figure}[t]
    \begin{center}
       \includegraphics[width=0.95\linewidth]{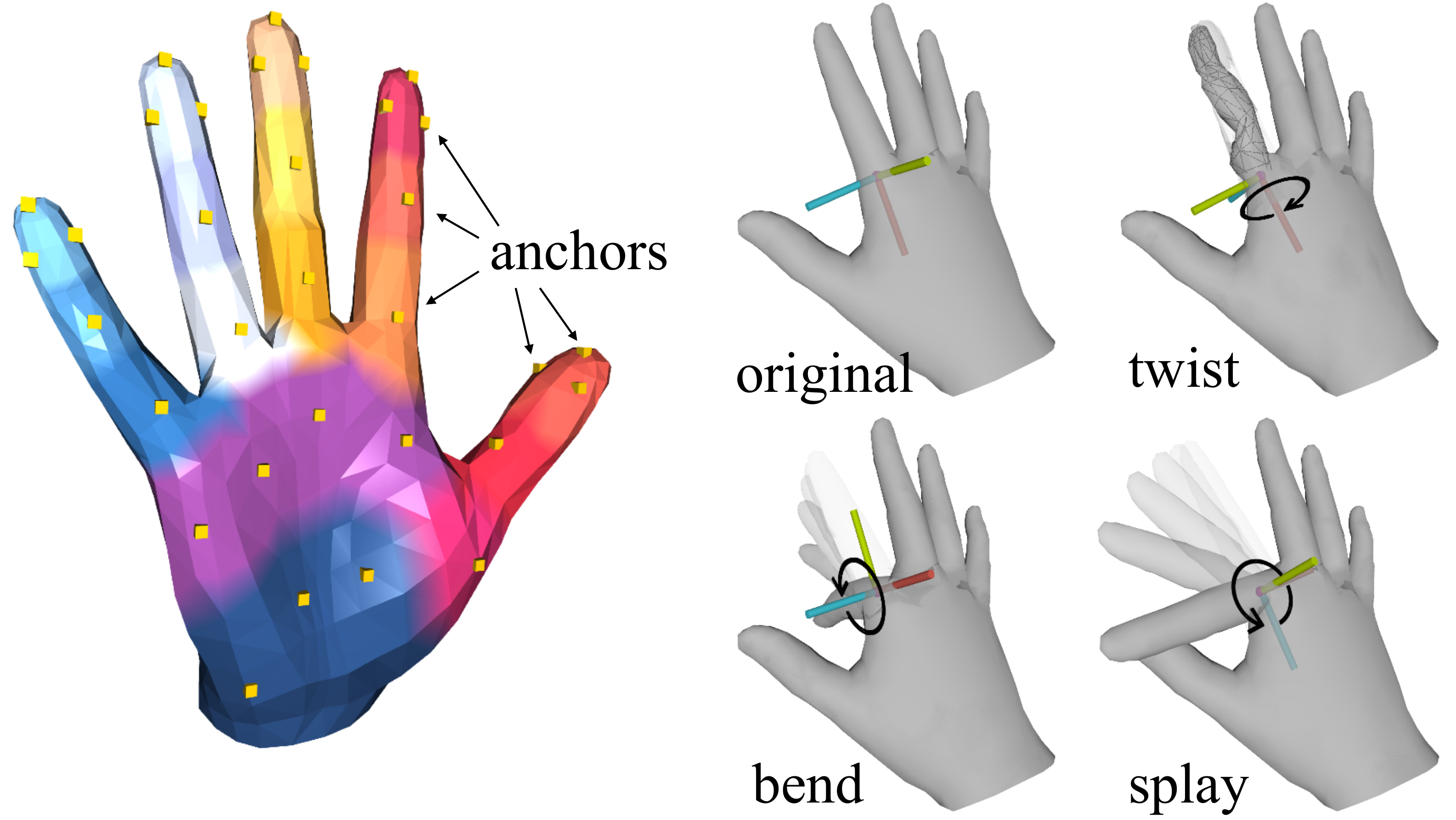}
    \end{center}
       \caption{\textbf{Illustration of the proposed A-MANO.} Left: the subdivision of hand regions and anchors attached to it. Right: the proposed \textit{twist-splay-bend} frame. }\vspace{-0.5em}
    \label{fig:handkin}
 \end{figure}

Another problem with the existing methods is the representation of the hand model.
Most researches adopted a skinning model, MANO \cite{romero2017embodied}, to represent hand.
MANO is considered to be flexible and deformable with its pose and shape parameters.
However, fitting on these high DoFs parameters is prone to anatomical abnormality. Researches in the robotics community adopted a dexterous hand \cite{kokic2020learning, garcia2020physics} in the off-the-shelf grasping software \cite{miller2004graspit}, which can almost guarantee a valid pose.
But the rigidity of those rod-like hand is less suitable for applications in CV$/$CG.
To make the best of both worlds, we propose a novel anatomically constrained hand model namely A-MANO (\S\ref{sec:k_mano}). It inherits the formulation of the skinning model and constrains the hand joints' rotation within a proposed \textit{twist-splay-bend} frame (Fig. \ref{fig:handkin}).


For evaluation, we report our scores on FHB \cite{garcia2018first} and HO3D \cite{hampali2020honnotate, hampali2019ho} dataset in terms of reconstruction and physical quality metrics. Note that, the ground truth of FHB is noisy and suffers from severe interpenetration \cite{karunratanakul2020grasping}. Since our method can avoid the penetration in the first place, our results are more visually and physically plausible. Therefore, we argue that, in this dataset, a higher reconstruction score does not necessarily benchmark the performance of the method.
While on HO3D, we achieve state-of-the-art performance on both reconstruction and physical metrics. The contributions of this paper are as follows.
\begin{itemize}
   \vspace{-0.5 em}\item We highlight contact in the hand-object interaction modeling task by proposing an explicit representation named CPF.
   \vspace{-0.5 em}\item We introduce A-MANO, a novel anatomical-constrained hand model that helps to mitigate pose's abnormality during optimization.
   \vspace{-0.5 em}\item We present a novel framework, MIHO, for modeling hand-object interaction. It can achieve state-of-the-art performance on several benchmarks.

\end{itemize}

\section{Related Work}
\vspace{0.2em}\noindent\textbf{3D Hand Reconstruction.}
Most of the existing 3D hand reconstruction methods \cite{boukhayma20193d, zhou2020monocular, baek2019pushing} adopted a parametric skinning hand, \eg MANO \cite{romero2017embodied} as a template.  To drive MANO, it is crucial to obtain joint rotation along hand kinematic tree. Boukhayma \etal \cite{boukhayma20193d} firstly proposed to regress the PCA components of the rotations.
Later, directly regressing the full rotations from 3D positions \cite{zhou2020monocular,yang2020bihand} has shown better performance. However, those high DoF regression is prone to pose abnormality. Thus, Spurr \etal \cite{spurr2020weakly} exploited biomechanical constraints over hand joints in training scheme. Different from \cite{spurr2020weakly}, we apply rotation constraints over the axes and angles in the proposed \textit{twist-splay-bend} coordinate frame.

\vspace{0.2em}\noindent\textbf{Hand-object Pose Estimation.}
In a wide range of topics in modeling hand-object interaction, the most commonly referred one is HO pose estimation \cite{hasson2019learning, hasson2020leveraging, doosti2020hope, gao2019graph, tekin2019h+}. In this regard, the earlier methods focused on either hand \cite{rogez20143d, romero2009monocular, tzionas2016capturing} or object \cite{tzionas20153d} pose alone, or estimated hand in grasping pose with knowing object shape prior \cite{feix2014analysis, cai2016understanding, cai2017ego, choi2017robust}. Jointly estimating hand and object pose was firstly presented by Romero \etal \cite{romero2010hands} via searching for nearest neighbors in a large database. Recently, learning-based frameworks have emerged in this area. Hasson \etal \cite{hasson2019learning,hasson2020leveraging} proposed two learning frameworks to recover hand-object meshes, one by synthesizing HO data under manipulation \cite{hasson2019learning} and the other by exploiting photometric consistency over video sequence \cite{hasson2020leveraging}. Doosti \etal \cite{doosti2020hope} employed the graph neural networks \cite{gao2019graph} to lift the 2D HO keypoints into 3D space. Tekin \etal \cite{tekin2019h+} adopted 3D YOLO \cite{redmon2017yolo9000} to predict HO pose in one stage. Korrawe \etal \cite{karunratanakul2020grasping} recovered HO model in a form of Signed Distance Function \cite{park2019deepsdf}.

\vspace{0.2em}\noindent\textbf{Contact Heuristic.}
Exploiting contact heuristic in hand-object interaction can be traced back to several decades before \cite{rijpkema1991computer, elkoura2003handrix, liu2009dextrous}.
Early works utilized some shape-specified contact physics (\eg cones and blocks \cite{rijpkema1991computer}) or predefined grasp \cite{liu2009dextrous} as prior.
Studies on capturing \cite{kry2006interaction} or imitating \cite{borst2005realistic} HO interaction also leveraged contact to satisfy the reality.
Later, the studies on grasping synthesis \cite{ye2012synthesis, garcia2020physics, kokic2020learning} and tracking \cite{oikonomidis2011full, kyriazis2013physically} turned to physical simulators to circumvent model's intersection.
Multi-point contact formulation was proposed in \cite{kim2015physics, holl2018efficient, antotsiou2018task}, which we found useful when applying physical constraints, \eg \cite{kim2015physics, holl2018efficient} used contact points to resolve penetration. For unified attraction and repulsion, most works employed heuristic such as proximity metric \cite{hasson2019learning, antotsiou2018task, tzionas2016capturing}, signed distance function \cite{karunratanakul2020grasping, brahmbhatt2019contactgrasp}, predefined contact pattern \cite {rogez2015understanding, brahmbhatt2019contactgrasp}, or turned to simulator \cite{kokic2019learning, kokic2020learning} for simplicity. Recently, Antotsiou \etal \cite{antotsiou2018task} refined the grasp by attracting fingers to its nearest point on object surface w.r.t  distance-based energy. Hasson \etal \cite{hasson2019learning} applied well-designed interaction losses which are also based on proximity metric. Although our method differs from all of the previous methods in terms of contact heuristic, we consider that both \cite{antotsiou2018task} and \cite{hasson2019learning} are still strong baselines. Thus we will compare our contact heuristic with theirs.

 \section{Anatomically Constrained A-MANO}
 \label{sec:k_mano}
 The proposed A-MANO inherits from a parametric skinning hand model, MANO \cite{romero2017embodied}, which drives an articulated hand mesh with pose parameters $\bm{\theta}$ and shape parameters $\bm{\beta}$. $\bm{\theta} \in \mathbb{R}^{15 \times 3}$ is 15 joint rotations along the hand kinematic tree. And $\bm{\beta} \in \mathbb{R}^{10}$
 represents the PCA components of hand shape. The main differences of A-MANO from MANO are: 1) the restriction on the joints' rotation axes and angles within the \textit{twist-splay-bend} frame; 2) the \textit{anchor} representation in the subdivision of hand region.

 \vspace{-0.8em}\paragraph{The \textit{Twist-splay-bend} Frame.} Fitting on 15 joint rotations of MANO requires high DoFs regression which may cause abnormal hand posture as shown in Fig. \ref{fig:monster}.
 Since the human hand can be modeled in a kinematic tree, and the majority of the joints only have one DoF about the \textit{bend} axis, we can impose constraints over the rotation about the unwanted axes. Therefore the proposed \textit{twist-splay-bend} Cartesian coordinate frame can be assigned to each joint along the kinematic tree. The frame' s $x,y,z$ axes are coaxial to the 3 revolute directions: \textit{twist}, \textit{splay}, and \textit{bend} direction on the basis of hand anatomy (Fig. \ref{fig:handkin} right). Then we can impose axial constraints in the \textit{twist} and \textit{splay} axes, and impose angular constraints w.r.t the \textit{bend} angle. Details of the \textit{twist-splay-bend} frame are elaborated in \textit{Supp} \ref{sec:coordinates}.


 \vspace{-0.8em}\paragraph{Anchors.} Since the hand mesh of different subjects are almost identical in the subdivision of hand region (\eg phalanges), we can interpolate several representative points (later we call it \textit{anchors}) on hand mesh to largely reduce the number of HO vertex pairs. Instead of attaching springs from object mesh to all the affinitive vertices on hand mesh, we only attach them on the several hand subregion centers, as we call it \textit{anchors} (Fig. \ref{fig:handkin} left). According to the statistics \cite{hasson2019learning, brahmbhatt2020contactpose} on the contact frequency of different hand parts, we first divide the full hand palm into 17 subregions: 3 for each phalange of 5 fingers, 1 for metacarpals, and another for carpals.
 Then, we interpolate up to 4 anchors for each subregion. We ignore all the vertices on the back side of the hand. Details of subregion division and anchors interpolation are described in \textit{Supp} \ref{sec:subregion}, \ref{sec:ancohor_selection}.

 \begin{figure*}
   \begin{center}
   \includegraphics[width=0.9\linewidth]{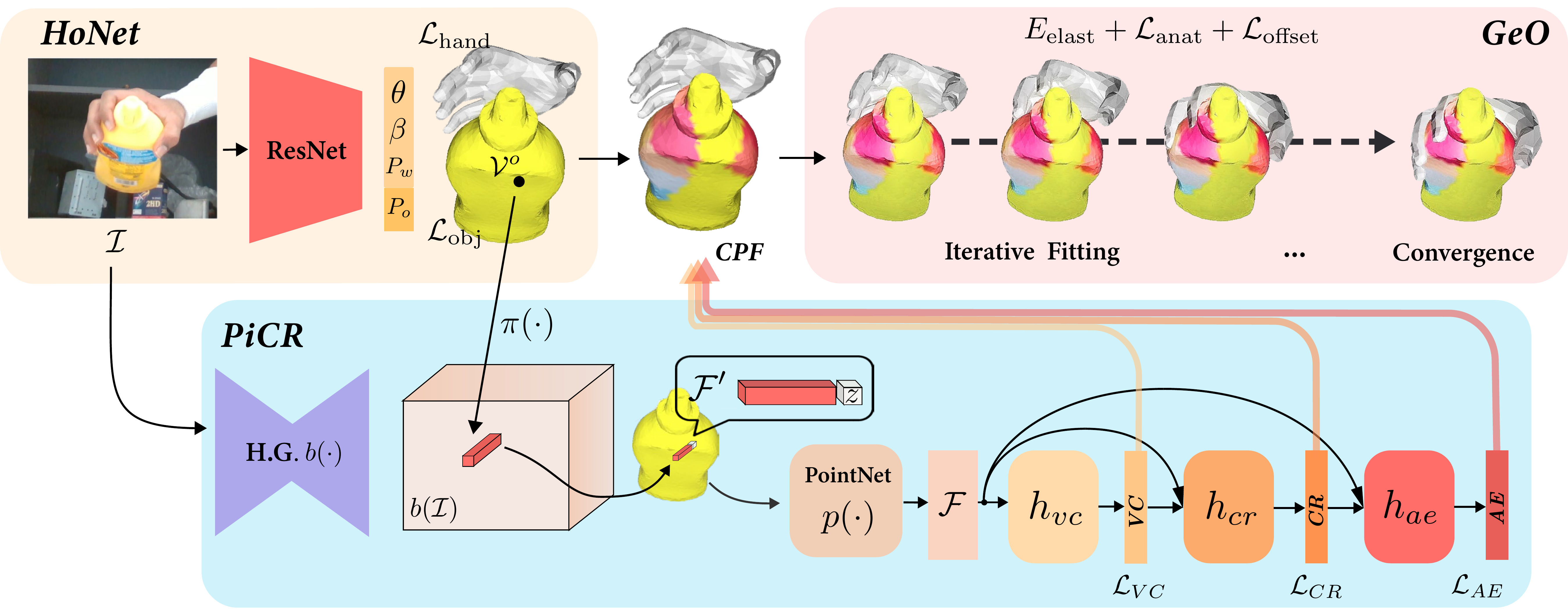}
   \end{center}
      \caption{\textbf{The architecture of the hybrid model MIHO.} The MIHO consists of three submodules: the first HONet estimates coarse poses of HO meshes, the second PiCR learns to recover the CPF and the last GeO retrieves the refined poses based on the CPF.}\vspace{-0.5em}
   \label{fig:pipeline}
\end{figure*}

 \section{Contact Potential Field}
 \label{sec:cpf}
\paragraph{Contact as Spring-Mass System.} A single contact is modeled as a spring-mass system which consists of a spring and two mass points on each side (hand and object). When the spring is at its rest position, it does not store energy, whilst it is stretched or compressed, according to Hooke's Law\footnote {\url{https://en.wikipedia.org/wiki/Hookes_law}}, it will store the elastic potential energy with the form: $\frac{1}{2}k|\bm{\Delta l}|^2$, where $k$ is the spring elasticity, and $|\bm{\Delta l}|$ is a certain ``\textit{distance}'' metric \wrt the spring's rest position.

In CPF, we define two types of spring: \textbf{\textit{attractive}} spring and \textbf{\textit{repulsive}} spring. The goal of \textit{attractive} spring is to pull the hand vertex $\bm{v}^h$ toward the object vertex $\bm{v}^o$ based on a given HO vertex pair affinity. And the goal of \textit{repulsive} spring is to push the $\bm{v}^h$ away from $\bm{v}^o$ along the $\bm{v}^o$ 's normal  if the $\bm{v}^h$ is in the vicinity of $\bm{v}^o$.
Apart from these definitions, we should also point out that the \textit{attractive} spring is bound with a certain pair of HO vertex affinity, while the \textit{repulsive} spring only takes effect in the neighborhood of HO vertex pairs at some point.

\vspace{0.3em}\textbf{\textit{- Attractive Spring.}}
We define the rest length of \textit{attractive} spring as $0$ in which the hand vertex and object vertex are in perfect contact,
and the \textit{distance} metric $|\bm{\Delta l}|$ as Euclidean distance. Given a HO affinity that includes a vertex pair: $\bm{v}^h_i$ and $\bm{v}^o_j$, the $|\bm{\Delta l}^{\rm{atr}}_{ij}|$ is equal to $\left \| \bm{v}^h_i - \bm{v}^o_j \right \|_2$. The  potential energy of the current \textit{attractive} spring is given by:
\begin{equation}
   \setlength\abovedisplayskip{5pt}
   E^{\rm{atr}}_{ij} = \frac{1}{2} k^{\rm{atr}}_{ij} * \left \| \bm{\Delta l}^{\rm{atr}}_{ij} \right \|_2^2
   \setlength\belowdisplayskip{5pt}
   \label{atr_spring}
\end{equation}


\vspace{0.3em}\textbf{\textit{- Repulsive Spring.}} We hope that the repulsion energy is high when $\bm{v}^h_i$ is penetrating or in the vicinity of $\bm{v}^o_j$, but gradually decays as the $\bm{v}^h_i$ moves away from the object, and finally becomes negligible at certain distance.
Given a proximate HO vertex pair: $\bm{v}^h_i$ and $\bm{v}^o_j$, We define a repulsive spring to model this behavior.
Supposing that the \textit{repulsive} spring has the rest position at $+\infty$ away along the object normal $\mathbf{n}^o_j$.  We adopt a heuristic \textit{distance} metric $|\bm{\Delta l}| = e^{-|\bm{\Delta l}^{\rm{rpl}}_{ij}|} - e^{-\infty} = e^{-|\bm{\Delta l}^{\rm{rpl}}_{ij}|}$, where $|\bm{\Delta l}^{\rm{rpl}}_{ij}| = (\bm{v}_i^h - \bm{v}_j^o) \cdot \mathbf{n}_j^o$ is the projection of the $(\bm{v}_i^h - \bm{v}_j^o)$ on the object normal $\mathbf{n}_{j}^{o}$. Thus, the potential energy of the current \textit{repulsive} spring is
 \begin{equation}
   \setlength\abovedisplayskip{5pt}
   \setlength\belowdisplayskip{5pt}
   E^{\rm{rpl}}_{ij} = \frac{1}{2} k^{\rm{rpl}}_{ij} * \big (e^{- |\bm{\Delta l}^{\rm{rpl}}_{ij}|} \big )^2
 \end{equation}
In literature, adopting repulsive effect along surface normal can be found in \cite{brahmbhatt2019contactgrasp, hampali2020honnotate}. \cite{hampali2020honnotate} (Eq. 10) also discussed that $e^{-(\cdot)}$ is an efficient heuristic concerning sub-sampled set of vertices.


 \vspace{-0.8em}\paragraph{Grasping inside Contact Potential Field.}
 By collecting all the \textit{attractive} and \textit{repulsive} springs, to form a natural grasp is equivalent to minimize the elastic energy:
 \begin{equation}
     \setlength\abovedisplayskip{5pt}
     \setlength\belowdisplayskip{5pt}
    E_{\rm{elast}}=\sum_{i}\sum_{j}(E^{\rm{atr}}_{ij} + E^{\rm{rpl}}_{ij})
    \label{rpl_spring}
 \end{equation}

As discussed in \S\ref{sec:k_mano}, the hand vertices can be simplified to subregion $anchors$, which will largely relax the difficulty of learning and fitting inside the CPF. Thus, for \textit{attractive} spring, we replace the $\bm{\Delta l}_{ij}$  in Eq.\ref{atr_spring} to $\bm{\Delta l}^\prime_{ij}=\bm{a}_i-\bm{v}^o_j$, where $\bm{a}_i$ is the closest anchor to $\bm{v}^h_i$. Besides, we would like to have the repulsion force be only applied to those HO affinity pairs that are of vertices in vicinity. Thus we set zero the repulsion energy when the vertex distance $\| \bm{v}_j^o - \bm{v}_i^h \|_2$ is greater than a threshold $t_{\rm{rpl}} = 20 \ mm$.

\vspace{-0.8em}\paragraph{Annotation of the \textit{Attractive} Springs ($k^{\rm{atr}}$).} While the attraction energy is bound with certain HO affinities, the repulsion energy is rather ambient and affinity-agnostic.
To integrate the CPF into learning framework, we only consider the $k_{ij}^{\rm{atr}}$ as the prediction of neural network.
To enable this, network shall have the abilities of 1) pairing the hand anchors and object vertices into HO affinity pair, \eg $(\bm{a}_i, \bm{v}^o_j)$; and 2) regressing the intensity of those affinity pairs, \eg $k_{ij}^{\rm{atr}}$. These require annotation of the \textit{attractive} springs $k^{\rm{atr}}$.

Given the ground-truth (\textit{gt.}) HO pose and their mesh model, we automatically annotate each $k_{ij}^{\rm{atr}}$ based on a heuristic of the $(\bm{a}_i, \bm{v}^o_j)$ pair distance. Since each $\bm{a}_i$ may be included in several affinity pairs, we hope the attraction energy stored in each spring at \textit{gt.} HO pose is well balanced. Thus we assign the \textit{gt.} $\hat{k}^{\rm{atr}}_{ij}$ a value that is inverse-proportional to the \textit{gt.} $|\bm{\Delta} \hat{\bm{l}}^{\rm{atr}}_{ij}|$. In order to train the network, we also bound the magnitude of $\hat{k}^{\rm{atr}}_{ij}$ by $0$ and $1$. Here we only provide a glimpse of the annotation heuristic of $\hat{k}^{\rm{atr}}_{ij}$:
\begin{equation}
   \setlength\abovedisplayskip{5pt}
   \setlength\belowdisplayskip{5pt}
   \hat{k}^{\rm{atr}}_{ij} = 0.5 * \cos \big (\frac{\pi}{s} * |\bm{\Delta} \hat{\bm{l}}^{\rm{atr}}_{ij} | \big ) + 0.5
\end{equation}
Empirically, we set the scale factor $s = 20 \ mm$ and reject those HO affinities with \textit{gt.} $|\bm{\Delta} \hat{\bm{l}}^{\rm{atr}}_{ij}| \ge 20 \ mm$.
As for the elasticity of \textit{repulsive} spring, we empirically set all ${k}_{ij}^{\rm{rpl}}$ to $1 \times 10^{-3}$. Detailed analysis of the \textit{gt.} $\hat{k}^{\rm{atr}}$ and the attraction-repulsion equilibrium are provided in \textit{Supp} \ref{sec:elast_energy_analysis}, \ref{sec:anchor_assignment},



\section{Hybrid Framework -- MIHO}\label{sec:approach}

With respect to the proposed CPF (\S\ref{sec:cpf}), our approach MIHO models the hand-object interaction in three stages, namely HoNet (\S\ref{honet}), PiCR (\S\ref{picr}), and GeO (\S\ref{geo}).

As shown in Fig.\ref{fig:pipeline}, firstly, given an RGB image $\mathcal{I}$,
HoNet predicts a coarse pose of  hand mesh $\mathcal{V}^{h} = \{\bm{v}_i^h \in \mathbb{R}^3 \ | \ i \le N_h \}$ and object mesh $\mathcal{V}^{o} = \{\bm{v}_j^o \in \mathbb{R}^3 \ | \ j \le N_o \}$, where $N_h$ and $N_o$ are the number of the vertex of hand and object respectively.
Then, PiCR learns to construct the CPF and collect the elastic energy $E_{\rm{elast}}$ in it.
Finally, GeO minimizes $E_{\rm{elast}}$ in CPF to yield the refined HO meshes ${^*\mathcal{V}}^{o}$, ${^*\mathcal{V}}^{h}$.

\subsection{Hand-object Pose Estimation Network, HoNet}\label{honet}
The HoNet first predicts coarse poses of HO meshes by the baseline model \textit{MeshRegNet} as in \cite{hasson2020leveraging}.
The outcomes from the baseline comprise in total 37 coefficients: object 6D pose $\bm{P}_o \in \mathfrak{se}(3)$ $(\mathbb{R}^6)$, hand wrist 6D pose $\bm{P}_w \in \mathfrak{se}(3)$, PCA components of MANO pose $\bm{\theta}_{\rm{pca}} \in \mathbb{R}^{15}$ and shape $\bm{\beta} \in \mathbb{R}^{10}$. With these coefficients, HoNet could place the HO meshes into camera frame. Details of the baseline can be referred to \cite{hasson2020leveraging}.

\subsection{Pixel-wise Contact Recovery Module, PiCR}\label{picr}
\begin{figure}[t]
\begin{center}
    \includegraphics[width=1.0\linewidth]{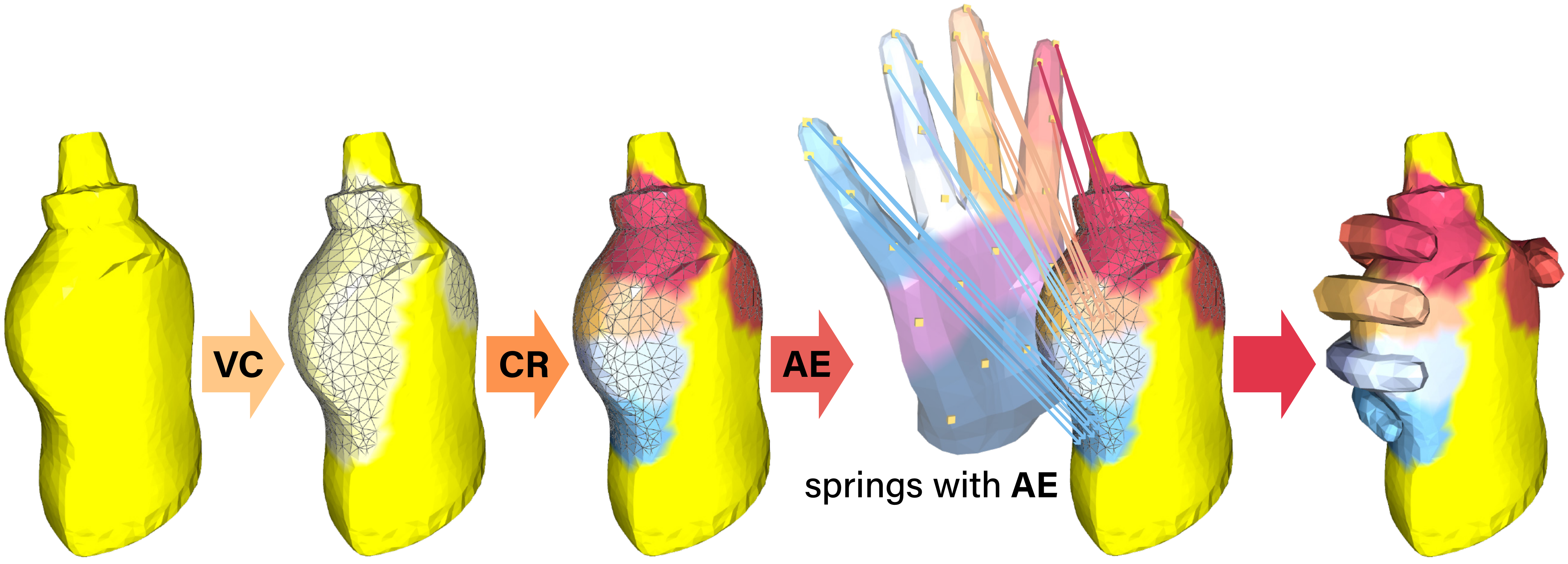}
\end{center}
    \caption{Illustration of assigning \textit{Vertex Contact}, \textit{Contact Region} and \textit{Anchor Elasticity} onto object surface.}
\label{fig:vccrae}
\end{figure}

With the coarse meshes of hand and object in HoNet, PiCR learns to recover the CPF by firstly paring the hand anchors and object vertices into HO affinity pairs and then regressing the spring elasticities that describe the affinities. To achieve this, PiCR yields three cascaded outcomes:
1) \textit{Vertex Contact} (\vC{}) decides which vertices on object are in contact with hand; 2) \textit{Contact Region} (\cR{}) decides the subregion that is most likely to contact with those vertices in \vC{}; 3) \textit{Anchor Elasticity} (\aE{}) represents the elasticities of the \textit{attractive} springs. With \vC, \cR, and \aE, we can then recover the CPF as illustrated in Fig. \ref{fig:vccrae}.

\setlength{\textfloatsep}{15pt}

\vspace{0.5em}\noindent\textbf{Vertex Contact.}
PiCR' s first outcome \vC{} $\in \mathbb{R}^{N_o}$ stands for the contact probability of object vertices. More specifically, \vC{}$[j]$ is a probability that implies the $j$-th object vertex $\bm{v}^o_j$ is in contact with hand.
The loss function of \vC{} is defined as a binary focal loss \cite{Lin2017focal}:
\begin{equation}
    \setlength\abovedisplayskip{5pt}
    \setlength\belowdisplayskip{5pt}
\mathcal{L}_{VC} = - \sum_j^{N_o} \mathds{1}^{img}_j * \alpha_j(1 - f_j)^{\gamma} \log(f_j)
\label{vertex_contact_loss}
\end{equation}
where $f_j = p_j$ if the \textit{gt.} $\hat{\bm{v}}^o_j$ belongs to any HO affinity, otherwise $f_j = (1 - p_j)$, and the $p_j$ is the predicted probability at \vC{}$[j]$. $\mathds{1}^{img}_j$ denotes whether the vertex $\bm{v}^o_j$ is projected inside the image. $\alpha_j$ is inverse class frequency and $\gamma$ is empirically set to $2$.

\vspace{0.5em}\noindent\textbf{Contact Region.}
PiCR's second outcome \cR{} $\in \mathbb{R}^{N_o \times 17}$ stands for the subregion probabilities of object vertices.
More specifically, for the $j$-th query, \cR{}$[j]$ contains $17$ probabilities that indicates $\bm{v}^o_j$ 's affinity toward $17$ hand subregions.
The loss function $\mathcal{L}_{CR}$ is defined as a multi-class focal loss.
\begin{equation}
    \setlength\abovedisplayskip{5pt}
    \setlength\belowdisplayskip{5pt}
\mathcal{L}_{CR} = - \sum_j^{N_o} \mathds{1}^{VC}_j * \mathds{1}^{img}_j * ( 1 - m_j)^{\gamma} \log(m_j)
\label{contact_region_loss}
\end{equation}
where the $m_j = \sum(p_j * t_j)$ in which $p_j =$ \cR{}$[j]$ $\in \mathbb{R}^{17}$ is the predicted per-subregion probabilities through \textit{softmax}, and $t_j \in \mathbb{R}^{17}$ is the \textit{gt.} subregion affinity of $\hat{\bm{v}}^o_j$ as a one-hot vector. $\mathds{1}^{VC}_j$ denotes that the \textit{gt.} \vC{} of $\hat{\bm{v}}^o_j$ is positive.

\vspace{0.5em}\noindent\textbf{Anchor Elasticity.} PiCR's third outcome \aE $\in \mathbb{R}^{N_o}$ stands for the predicted elasticity of \textit{attractive} springs $k^{\rm{atr}}$. More specifically, $\aE{}[j]$ is the elasticity $k_{ij}^{\rm{atr}}$ of an \textit{attractive} spring that connects $\bm{v}^o_j$ to its affinitive anchor $\bm{a}_i$ in the predicted subregion: $argmax(\cR{}[j])$.
The loss function $\mathcal{L}_{AE}$ is defined as a binary cross-entropy (BCE):
\begin{equation}
    \setlength\abovedisplayskip{5pt}
    \setlength\belowdisplayskip{5pt}
    \vspace{-0.3em}
\mathcal{L}_{AE} = \sum_j^{N_o} \mathds{1}^{VC}_j * \mathds{1}^{img}_j * \textit{BCE}(\: k_{ij}^{\rm{atr}}, \: \hat{k}_{ij}^{\rm{atr}})
\label{anchor_elasti_loss}
\end{equation}
where the $\hat{k}_{ij}^{\rm{atr}}$ is the \textit{gt.} elasticity described in \S\ref{sec:cpf}.

With the predicted \vC{}, \cR{} and \aE{}, as well as the coarse meshes $\mathcal{V}^{o}, \mathcal{V}^{h}$ in HoNet,
PiCR finally recovers the CPF and collects the elastic energy $E_{\rm{elast}}$ as described in Algm.\ref{alg:construct-potential-field}. We empirically set the probability threshold of \vC{}: $t_{\rm{vc}} = 0.8 $ and the distance threshold: $t_{\rm{rpl}} = 20\ mm$.

\setlength{\textfloatsep}{5pt}
\begin{algorithm}[t] \label{alg:construct-potential-field}
\begin{small}
\caption{Procedure of recovering CPF}
\KwIn{$\mathcal{V}^o, \mathcal{V}^h, \vC{}, \cR{}, \aE{}$}
\KwOut{$E_{\rm{elast}}$: elastic energy}
recovery anchors: $\mathcal{A} \leftarrow linear\_interpolation(\mathcal{V}^h) $\;
\ForEach{$j \in \{j \ |\ j \le N_o, \vC{}[j] > t_{\rm{vc}}\}$}
{
    recover subregion id: $r \leftarrow argmax(\cR{}[j])$\;
    \ForEach{$\bm{a}_i \in \mathcal{A}_r$ \textup{ (anchors in subregion $r$)}}
    {
        recover elasticity: $k_{ij}^{\rm{atr}} \leftarrow \aE{}[j]$\;
        $E_{\rm{elast}}  \ +\leftarrow \frac{1}{2} * k_{ij}^{\rm{atr}} \big \| \bm{a}_i -  \bm{v}_j^o \big \|_2^2 $\;
    }
    \ForEach{ $i \in \{ i\ |\ i \le N_h, \big \| \bm{v}_i^h -  \bm{v}_j^o \big \|_2^2 \le t_{\rm{rpl}} \}$ }{
        $E_{\rm{elast}} \ +\leftarrow \frac{1}{2} * k_{ij}^{\rm{rpl}} \big | \exp (- (\bm{v}_i^h - \bm{v}_j^o) \cdot \mathbf{n}_j^o  ) \big |^2$\;

    }
}
\end{small}
\end{algorithm}

\vspace{0.5em}\noindent\textbf{PiCR's Framework.} The proposed PiCR consists of a backbone $b$ that extracts features from image, an encoder $p$ that converts image features to object vertex features, and 3 heads $h_{\rm{vc}}$, $h_{\rm{cr}}$ and $h_{\rm{ae}}$ which sequentially convert those features into \vC{}, \cR{}, and \aE{}.
As illustrated in Fig. \ref{fig:pipeline}, the process of feature extraction in PiCR can be expressed as:
\begin{equation}
    \setlength\abovedisplayskip{8pt}
    \setlength\belowdisplayskip{8pt}
\mathcal{F}^\prime = \Big [ f \big ( \pi(\mathcal{V}^o), b(\mathcal{I}) \big ),  z(\mathcal{V}^o)  \Big ]; \quad  \mathcal{F}=p(\mathcal{F}^\prime)
\label{picr function}
\end{equation}
where $b(\cdot)$ is the hourglass networks \cite{newell2016stacked}, $\pi(\cdot)$ is the perspective camera projection, and $f(\cdot)$ stands for aligning $\mathcal{V}^o$ 's 2D projection $\pi(\mathcal{V}^o)$  with the image features $b(\mathcal{I})$ through bilinear sampling. Inspired from Eq.(1) in \cite{saito2019pifu}, we also append the object's root-relative $z$ value $z(\mathcal{V}^o)$ at the end of $f(\cdot)$ to form the pixel-wise features $\mathcal{F}^\prime$.
Next, a PointNet \cite{qi2017pointnet} encoder $p(\cdot)$ is adopted to convert $\mathcal{F}^\prime$ to its point-wise features $\mathcal{F}$.

The process of three PiCR's heads can be expressed as:
\begin{equation}
    \setlength\abovedisplayskip{8pt}
    \setlength\belowdisplayskip{8pt}
    \vC{} = h_{\rm{vc}}(\mathcal{F});\ \cR{} = h_{\rm{cr}}(\vC{}, \mathcal{F});\ \aE{} = h_{\rm{ae}}(\cR{}, \mathcal{F})
\label{picr function 2}
\end{equation}
where all the heads are presented as multi-layer perceptrons. We provide implementation details in \textit{Supp} \ref{sec:more_impl_details}.

\subsection{Grasping Energy Optimizer, GeO}
\label{geo}
The fitting part: Grasping Energy Optimizer (GeO) aims to refine the HO pose \wrt the recovered CPF.
For the object part, we adjust its 6D pose $\bm{P}_o \in \mathfrak{se}(3)$.
For the hand part, we jointly adjust the A-MANO' s 15 joint rotations $\{ \bm{R}_{j} \in \mathfrak{so}(3)\ |\ j \le 15 \}$ and a wrist pose $\bm{P}_w \in \mathfrak{se}(3)$.

In order to mitigate the abnormal hand posture during optimization, we also define an anatomical cost function $\mathcal{L}_{\rm{anat}}$ that penalizes the unwanted axial components and angular values of the $15$ rotations in the proposed \textit{twist-splay-bend} coordinate frame.
First, for the joints along hand kinematic tree, we penalize the component of rotation axis $\mathbf{a}^{rot}$ on \textit{twist} direction: $\mathbf{n}^{twist}$, since any component that causes the finger twisting along its pointing direction is prohibited. Second, for the joints that do not belongs to 5 knuckles, we also penalize the component of $\mathbf{a}^{rot}$ on \textit{splay} direction: $\mathbf{n}^{splay}$. Last, we penalize the rotation angle $\phi^{bend}$ that revolves about the \textit{bend} axis if it is greater than $\pi/2$. The total anatomical cost can be written as:
\begin{equation}
\setlength\abovedisplayskip{8pt}
\setlength\belowdisplayskip{5pt}
\begin{small}
\begin{aligned}
\mathcal{L}_{\rm{anat}} = &\sum_{j \in {\rm{all}}}\mathbf{a}^{rot}_{j} \cdot \mathbf{n}^{twist}_{j} + \sum_{j \notin \rm{knuck}}{\mathbf{a}_j^{rot} \cdot \mathbf{n}_j^{splay}}\\
+ &\sum_{j \in \rm{all}} \max \left ((\phi_j^{bend} - \frac{\pi}{2}), 0 \right )
\label{hand_loss}
\end{aligned}
\end{small}
\end{equation}
We also penalize the offset of the refined hand-object vertices $^*\mathcal{V}^{o}$, $^*\mathcal{V}^{h}$ from their initial estimation ${\mathcal{V}}^{h}$, ${\mathcal{V}}^{o}$ in form of \textit{l2} distance: $\mathcal{L}_{\rm{offset}}$. We implement GeO in PyTorch with Adam solver. The whole optimization process can be expressed as:
\begin{equation}
\begin{aligned}
    {^*\mathcal{V}}^{o}, {^*\mathcal{V}}^{h} \longleftarrow  \mathop{argmin}_{\bm{P}_o, \bm{P}_w, \bm{R}_j} (E_{\rm{elast}} + \mathcal{L}_{\rm{anat}} + \mathcal{L}_{\rm{offset}})
\label{offset_loss}
\end{aligned}
\end{equation}

\begin{table*}
    \begin{center}
    \begin{footnotesize}
    \resizebox{\textwidth}{!}
    {%
        \begin{tabular}{l|ccccc|cccc|cc}
        \toprule
        Datasets & \multicolumn{5}{c|}{FHB} & \multicolumn{4}{c|}{HO3Dv1$^+$} & \multicolumn{2}{c}{HO3Dv2$^-$} \\
        \cmidrule(lr){1-1} \cmidrule(lr){2-6} \cmidrule(lr){7-10} \cmidrule(lr){11-12}
        Method & Ours$^{\dag}$ & Ours$^{\ddag}$  &  \textit{gt.} &  \cite{hasson2020leveraging}  & ObMan$^*$  & Ours$^{\dag}$ & Ours$^{\ddag}$ & \textit{gt.} & \cite{hasson2020leveraging}$^+$  & Ours$^{\ddag}$ &  \cite{hasson2020leveraging} \\
        \midrule
        \cellcolor{White} Hand MPVPE ($mm$) $\downarrow$ & \cellcolor{White}21.16 & \cellcolor{White}19.54 & \cellcolor{White}0 & \cellcolor{White}\textbf{17.51} & \cellcolor{White}18.42 & \cellcolor{White}24.56 & \cellcolor{White}\textbf{23.99} & \cellcolor{White}0 & \cellcolor{White}24.80 & \cellcolor{White}- & \cellcolor{White}- \\
        \cellcolor{Gray} Object MPVPE ($mm$) $\downarrow$ & \cellcolor{Gray}\textbf{21.06} & \cellcolor{Gray}21.57  & \cellcolor{Gray}0 & \cellcolor{Gray}\textbf{21.06} & \cellcolor{Gray}21.17 & \cellcolor{Gray} \textbf{18.10} & \cellcolor{Gray}19.15 & \cellcolor{Gray}0 & \cellcolor{Gray}\textbf{18.10} & \cellcolor{Gray}\textbf{73.28} $^{\diamondsuit}$ & \cellcolor{Gray}75.77 $^{\diamondsuit}$ \\
        \cellcolor{White} Penetra. depth ($mm$) $\downarrow$ & \cellcolor{White}\textbf{16.13} & \cellcolor{White}16.92 & \cellcolor{White}19.55 & \cellcolor{White}20.63 & \cellcolor{White}19.76 & \cellcolor{White}11.87 & \cellcolor{White}\textbf{11.42} & \cellcolor{White}7.55 & \cellcolor{White}18.57 & \cellcolor{White}\textbf{16.47}  & \cellcolor{White}20.02 \\
        \cellcolor{Gray} Solid intersec. vol. ($cm^3$) $\downarrow$ & \cellcolor{Gray}12.56 & \cellcolor{Gray}\textbf{11.76} & \cellcolor{Gray}20.41 & \cellcolor{Gray}21.10 & \cellcolor{Gray}16.16 & \cellcolor{Gray}3.63 & \cellcolor{Gray}\textbf{3.46} & \cellcolor{Gray}3.57  & \cellcolor{Gray}9.62 & \cellcolor{Gray}\textbf{7.44} & \cellcolor{Gray}9.25 \\
        \cellcolor{White} Disjoint. distance ($mm$) $\downarrow$ & \cellcolor{White}24.54 & \cellcolor{White}\textbf{22.41} & \cellcolor{White}37.28 & \cellcolor{White}37.40 & \cellcolor{White}27.95 & \cellcolor{White}\textbf{11.71} & \cellcolor{White}11.83 & \cellcolor{White}14.53 & \cellcolor{White}18.62 & \cellcolor{White}\textbf{37.04} & \cellcolor{White}41.41 \\
        \cellcolor{Gray} Displacement ($mm$) $\downarrow$ & \cellcolor{Gray}58.79 & \cellcolor{Gray}\textbf{58.02} & \cellcolor{Gray}63.40 & \cellcolor{Gray}65.48 & \cellcolor{Gray}59.41 & \cellcolor{Gray}28.16 & \cellcolor{Gray}27.66 & \cellcolor{Gray}12.37 & \cellcolor{Gray}\textbf{25.68} & \cellcolor{Gray}\textbf{39.33} & \cellcolor{Gray}41.03 \\
        \bottomrule
        \end{tabular}
    }
    \end{footnotesize}
    \end{center}
    \caption{\textbf{Quantitative results and detailed comparison with the previous state-of-the-art \cite{hasson2020leveraging, hasson2019learning} on the FHB and HO3D datasets.} ``\textit{gt.}'' denotes the ground-truth.
    ``$\dag$'' denotes ours \textit{hand-alone} optimization setting and ``$\ddag$'' denotes the jointly \textit{hand-object} setting.
    ``$*$'' denotes the reproduced ObMan \cite{hasson2019learning}. ``$\diamondsuit$'' denotes the \textit{wrist-relative} object vertex error.  ``-'' indicates the results that are not available.}\vspace{-0.5em}
    \label{tab:benchmark}
\end{table*}

 \begin{figure*}
   \begin{center}
   \includegraphics[width=0.92\linewidth]{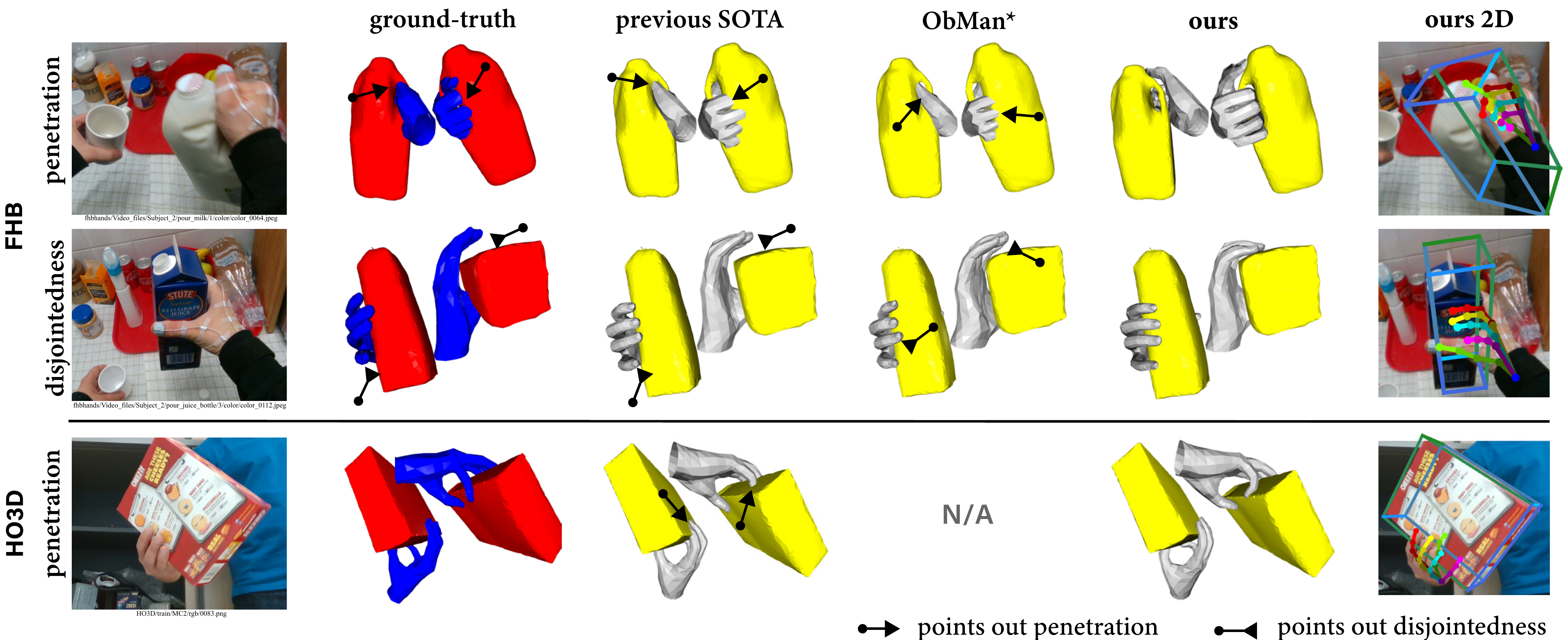}
   \vspace{-0.8em}
   \end{center}
      \caption{\textbf{Qualitative Comparison with ground-truth and  previous arts on the FHB and HO3D datasets.}}\vspace{-0.8em}
   \label{fig:qualitative_comparison}
\end{figure*}

\vspace{-0.8em}\section{Experiments and Results}


\subsection{Datasets}
\label{sec:dataset}
We would like to train and evaluate MIHO \wrt the real-world dataset that involves human hand interacting with textured object.
In the community, there exist mainly four datasets that contain images and ground-truth 3D HO annotation, namely ObMan \cite{hasson2019learning}, FHB \cite{garcia2018first} and HO3D \cite{hampali2019ho, hampali2020honnotate} and  ContactPose \cite{brahmbhatt2020contactpose}. However, only FHB and HO3D satisfy our requirements in this study.

\vspace{-0.8em}\paragraph{First-person Hand Action Benchmark, FHB.} FHB is a first-person RGBD video dataset of hand in manipulation with objects. The ground-truth of hand poses was captured via magnetic sensors. In our experiments, we use a subset of FHB that contains 4 objects with a scanned model and pose annotation. We adopt the \textit{action} split following the protocol given by \cite{hasson2020leveraging,tekin2019h+}, and filter out the samples with a minimum HO distance greater than 5 $mm$, which yields us 7223 samples for training and 7373 for testing.

\vspace{-0.8em}\paragraph{HO3D.}
\label{ho3d}
HO3D is another dataset that contains precise hand-object pose during the interaction. Due to historical reasons, there is two versions of HO3D, namely v1 \cite{hampali2019ho} and v2 \cite{hampali2020honnotate}. In our experiments, we mainly compare our methods with the baseline \cite{hasson2020leveraging} on HO3Dv1, but also conduct several comparisons with the recently released pre-trained model of \cite{hasson2020leveraging} on HO3Dv2. Similar to FHB, we filter out samples with distance threshold 5$mm$. It's also worth mentioning that, since our method requires a known object model, as well as a stable grasping configuration, nearly 5448 samples in HO3Dv2 test set are not suitable for our methods to report. Therefore, we manually select 6076 samples in HO3dv2 test set to compare MIHO with \cite{hasson2020leveraging}. We call this split by HO3Dv2$^-$. Besides, training HO3Dv1 in previous methods \cite{hampali2019ho, hasson2020leveraging} requires an extra synthetic dataset that is not publicly available. Thus we manually augment the HO3Dv1 training set (referred as HO3Dv1$^+$) and reproduce the results (referred as \cite{hasson2020leveraging}$^+$) comparable with those in \cite{hasson2020leveraging}. Details of HO3Dv2$^-$ selection and the augmentation procedures are provided in \textit{Supp} \ref{sec:ho3d_selection}, \ref{sec:ho3d_data_augmentation}.


\subsection{Metrics}
\label{sec:metrics}
Modeling the HO interaction requires not only a proper pose of both hand and object but also a natural grasp configuration. Here, we report 5 metrics in total that cover both reconstruction and grasp quality. Note that, since considering either of those metrics alone may yield misleading comparison, we consider them \textbf{together} for evaluation.


\noindent\textbf{MPVPE.} We compute the mean per vertex position error for both hand and object in camera space to assess the quality of pose estimation.

\noindent\textbf{Penetration Depth (PD).} To measure how deep that the hand is penetrating the object's surface, we calculate the penetration depth that is the maximum distance of all the penetrated hand vertices to their closest object surface.

\noindent\textbf{Solid Intersection Volume (SIV).} To measure how much space intersection that occurs during estimation, we voxelize the object mesh into $80^3$ voxels, and calculate the sum of the voxel volume inside the hand surface.

\noindent\textbf{Disjointedness Distance (DD).} We also encourage stable HO contact, which can be depicted as attracting fingertips onto the object surface.  Therefore, we define the disjointedness metrics as the average distance of hand vertices in 5 fingertips region to their closet object surface.

\noindent\textbf{Simulation Displacement (SD).} We further evaluate the grasp stability in a modern physics simulator \cite{coumans2017pybullet}. We measure the average displacement of object's center over a fixed time period by holding the hand steadily and applying gravity to the object \cite{hasson2019learning}.

\subsection{Comparison with State-of-the-Arts}
For the FHB dataset, we compare our methods with the previous SOTA \cite{hasson2020leveraging, hasson2019learning} of hand-object reconstruction. For  \cite{hasson2020leveraging}, we select the results under the setting of full data supervision. Since \cite{hasson2020leveraging} didn't exploit any repulsion and attraction loss during training, direct comparisons on intersection and disjointedness may not be convincing enough. While the contact losses were considered in another work named ObMan \cite{hasson2019learning}, it only represented the genus 0 object mesh as a deformable icosphere, which is also not directly comparable with ours (known object model). To ensure rational comparisons, we migrate the \textit{repulsion loss} and \textit{attraction loss} from ObMan to the \textit{MeshRegNet} in \cite{hasson2020leveraging}, and reproduce the results on par with it. We call this adaptation: ObMan$^*$. For the HO3Dv1 dataset, we compare our results with the reproduced \cite{hasson2020leveraging}$^+$.

We report our results under two experimental settings: 1) \textbf{\textit{hand-alone}} that fixes the object at the initial prediction in HoNet, and only optimizes the hand pose in GeO; 2) \textbf{\textit{hand-object}} that jointly optimizes the hand and object poses in GeO. In Tab.\ref{tab:benchmark} we show our comparisons with the previous SOTA in all 5 metrics. For FHB dataset, as analyzed in \cite{brahmbhatt2020contactpose}, its ground-truth suffers from frequent interpenetration. We find that lower vertex error does not necessarily benchmark a higher reconstruction quality. Indeed, as shown in Tab.\ref{tab:benchmark} (col. 4, 5), either ground-truth or \cite{hasson2020leveraging} reveals substantial solid intersection volume, penetration depth and disjointedness. We find that MIHO outperforms \cite{hasson2020leveraging} by a margin of 3.71 $mm$ in penetration depth, 9.34 $cm^3$ in solid intersection volume, and 14.99 $mm$ in disjointedness distance, while only suffers from minor performance cost in hand MPVPE of 2.03 $mm$ and object MPVPE of 0.51 $mm$. In the mean time, our simulation displacement also demonstrates the stability of our predicted grasp.
These are consistent with our expectation that the CPF can by nature repulse intersection away and attract disjointedness to touch.
As for HO3Dv1 testing set, our method also outperformed the previous SOTA over the most metrics. In terms fo simulation displacement, we found \cite{hasson2020leveraging}$^+$ slightly outperforms us by 1.98 $mm$. Based on our inspection in the Bullet \cite{coumans2017pybullet} simulator, their stability are mainly attributed to the forces resulting from the intersection that balance each other.  Visual comparisons are shown in Fig. \ref{fig:qualitative_comparison}.
As for HO3Dv2, since we only test MIHO on the subset: HO3Dv2$^-$, our results are not directly suitable for submitting to its online evaluation server. Thus, we only report the object 3D vertex errors on HO3Dv2$^-$ based on the given annotation. We firstly align the predicted object vertex to the predicted hand wrist joint, then compute the \textit{wrist-relative} object vertex error with those in ground-truth. Detailed comparisons are in Tab. \ref{tab:benchmark} (col. 11, 12).


\begin{figure}
    \begin{center}
       \includegraphics[width=1.0 \linewidth]{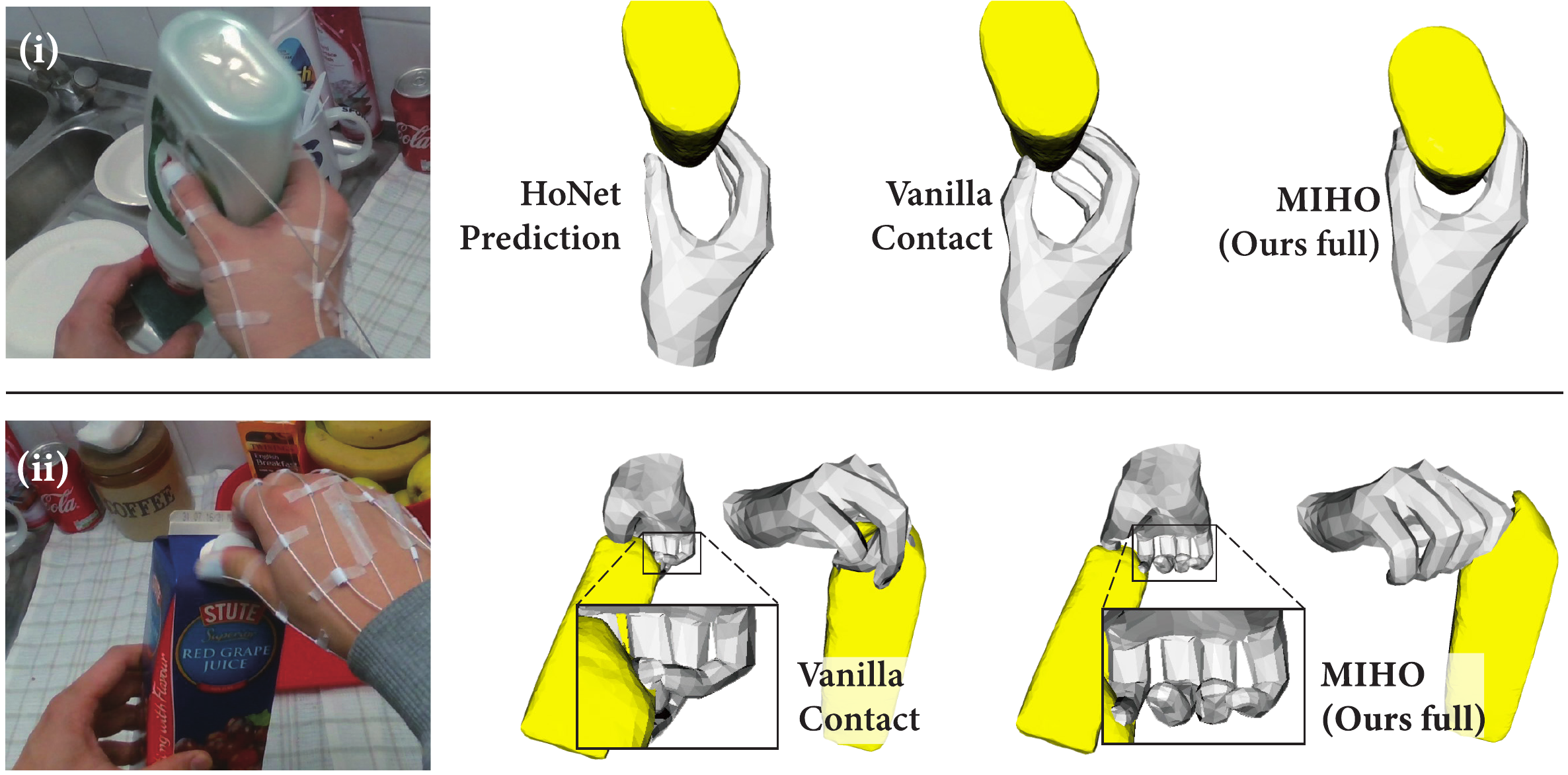}
    \end{center}
    \vspace{-0.5em}\caption{Comparisons of MIHO with simple contact heuristic.}
    \label{fig:heuristic}
 \end{figure}

\begin{figure}
    \begin{center}
       \includegraphics[width=0.9 \linewidth]{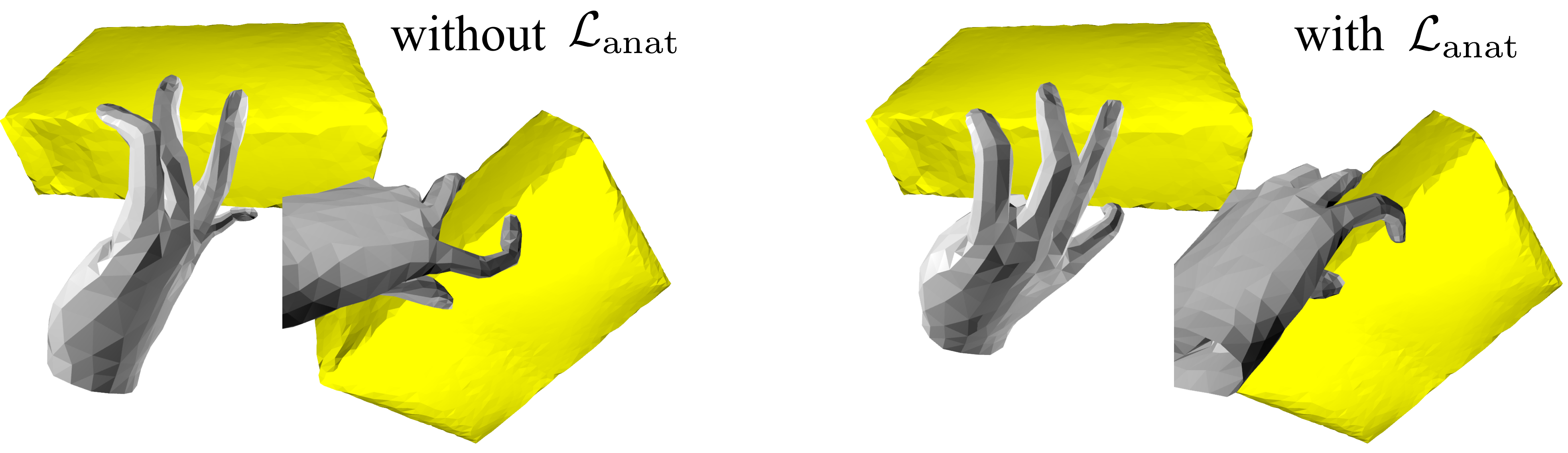}
    \end{center}
    \vspace{-0.5em}\caption{Example to illustrates the efficacy of our proposed A-MANO with anatomical constraints ($\mathcal{L}_{\rm{anat}}$).} \vspace{0.5em}
    \label{fig:monster}
 \end{figure}

\subsection{Ablation Study}
\label{sec:ablation}
In this experiment, we further evaluate the effectiveness of the proposed CPF and A-MANO. In the main text, we include three of the most representative studies. The ablation studies are mainly conducted on the FHB test set with \textit{action} split.  For more studies on 1) impact from the magnitude of $k^{\rm{rpl}}$; 2) A-MANO with PCA pose; 3) unwanted twist correction; please visit \textit{Supp} \ref{sec:more_ablation_study}.

\vspace{0.8em}\noindent\textbf{Comparison with simple Distance-based Contact Heuristics. } To show the superiority of the CPF over the distance-based contact heuristics, we compare the fitting stage of MIHO with two simple yet strong baselines:
(a) \textit{Vanilla Contact} that removes the $E_{\rm{elast}}$ term in Eq. \ref{offset_loss} and purely attracts the anchors on fingertips to its nearest object vertex (similar to \cite{antotsiou2018task}) in a given threshold which we set as 20 $mm$; (b) \textit{ObMan Contact} that replaces $E_{\rm{elast}}$ in Eq. \ref{offset_loss} by the well-designed interaction losses in ObMan \cite{hasson2019learning}. All the three experiments start from the same HO pose predicted by HoNet (\S\ref{honet}). We show in Tab. \ref{table:ablation_contact_heuristic}  that by exploiting CPF, MIHO can surpass the simple baselines on most of the metrics. Note that, since both (a) and (b) directly optimize the disjointedness term, their results show better resistance on it.
The last column in Tab. \ref{table:ablation_contact_heuristic} shows that our methods can save average time per iteration by 46\% compared with \textit{ObMan Contact}.
We also conduct two qualitative comparisons in Fig. \ref{fig:heuristic}. The first one shows that CPF can learn the contact \textit{semantics} to guide the optimization that better matches visual cues, whereas the \textit{Vanilla Contact} fails to form a valid grasp. The second shows that CPF can maintain subtle interaction, as no attraction will be applied on those non-affinitive vertex pairs (see ring and pinky fingers when unscrewing the juice cap).

\vspace{0.8em}\noindent\textbf{Effectiveness of \textit{Repulsive} Springs.} To measure the efficacy of \textit{repulsive} springs in CPF, we remove all the repulsion energy $E^{\rm{rpl}}$ induced by them, leaving the attraction as the unique type of energy applied on hand and object. As we expected, the result in Tab. \ref{table:ablation} witnesses the accumulation of PD and SIV.
To note, even without the \textit{repulsive} springs, we still witness a remarkable improvement of PD and SIV over the FHB ground-truth. This is attributed to the repulsive behavior of the \textit{attractive} springs: when hand is inside the object surface, the energy stored in the \textit{attractive} springs will act as repulsion that pushes out the hand.

\vspace{0.8em}\noindent\textbf{Effectiveness of the Anatomical Constraints.} We further highlight the efficacy of adopting the anatomical constraints. We conduct a contrastive experiment whose only difference is the absence of $\mathcal{L}_{\rm{anat}}$. Both experiments start from a zero (flat) hand and minimize the $E_{\rm{elast}}$ based on the same predicted CPF. We show in Fig. \ref{fig:monster} that the anatomical constraints are able to effectively prevent abnormality during the optimization.

\begin{table}
    \renewcommand{\arraystretch}{1}
    \begin{center}
    \resizebox{\linewidth}{!}{
        \begin{tabular}{l|ccccc|c}
        \toprule
        \multirow{2}{*}{Settings} & \multicolumn{5}{c|}{{Scores}} & \multirow{2}{*}{$t_{\rm{iter}}$($ms$)} \\
        \cmidrule(r){2-6}
         & HE $\downarrow$ & OE  $\downarrow$ & PD $\downarrow$ & SIV  $\downarrow$ & DD $\downarrow$ \\
        \midrule
        \textbf{MIHO} (ours full) & 19.54 & 21.57 & 16.92 & 11.76 & 22.41 & 55.77 \\
        (a) Vanilla Contact & 24.01 & 24.29 & 18.36 & 15.64 & 16.32 & 45.40 \\
        (b) ObMan Contact & 22.15 & 22.54 & 15.13 & 16.20 & 11.97 & 103.41 \\
        \bottomrule
        \end{tabular}
    }
    \end{center}
    \caption{Ablation study on the different contact heuristics. HE, OE stands for 3D hand and object vertex error. PD, SIV and DD are the abbreviation of metrics in \S\ref{sec:metrics}}
    \label{table:ablation_contact_heuristic}
\end{table}

\begin{table}
    \footnotesize
    \renewcommand{\arraystretch}{1}
    \begin{center}
        \begin{tabular}{l|ccc}
        \toprule
         Settings & PD$\downarrow$ & SIV$\downarrow$ & DD$\downarrow$\\
        \midrule
        \textbf{with $E^{\rm{rpl}}$} (ours full) & 16.92& 11.76& 22.41 \\
        without $E^{\rm{rpl}}$ & 17.79 & 13.76 & 20.27\\
        \textit{gt.} FHB  & 19.55 & 20.41 & 37.28 \\
        \bottomrule
        \end{tabular}
    \end{center}
    \caption{Ablation study on the \textit{repulsive} springs.}\vspace{0.5em}
    \label{table:ablation}
\end{table}

\section{Conclusion}
In this work, we propose a novel contact representation named CPF and a learning-fitting hybrid framework MIHO to help modeling hand and object interaction.  Comprehensive evaluations show that our methods, while being able to recover precise hand-object pose, can also effectively 1) avoid interpenetration and control disjointedness, and 2) prevent abnormality in hand pose. We hope CPF can serve as an effective contact representation for future works on hand-object interaction.
Later, we also plan to develop for an object-agnostic representation of CPF, for the interaction in general cases.


{\small
\bibliographystyle{ieee_fullname}
\bibliography{egbib}
}

\clearpage
\section*{Appendix}
\label{sec:appdx}

In the supplemental document, we provide:
\begin{itemize}
\setlength{\itemsep}{0pt}
    \item [\S\ref{suppsec:a-mano}] Anatomically Constrained A-MANO.
    \item [\S\ref{suppsec:springs}] Detailed Analysis of the Spring's Elasticity.
    \item [\S\ref{suppsec:ho3d}] Detailed Analysis of the HO3D Dataset.
    \item [\S\ref{suppsec:expm}] More Experiments and Results.
    \item [\S\ref{suppsec:morequal}] More Qualitative Results.
\end{itemize}

\begin{appendix}

\section{Anatomically Constrained A-MANO}\label{suppsec:a-mano}
\subsection{Derivation of \textbf{\textit{Twist-splay-bend}} Frame.}\label{sec:coordinates}

\begin{figure}
   \centering
   \includegraphics[width=0.9\linewidth]{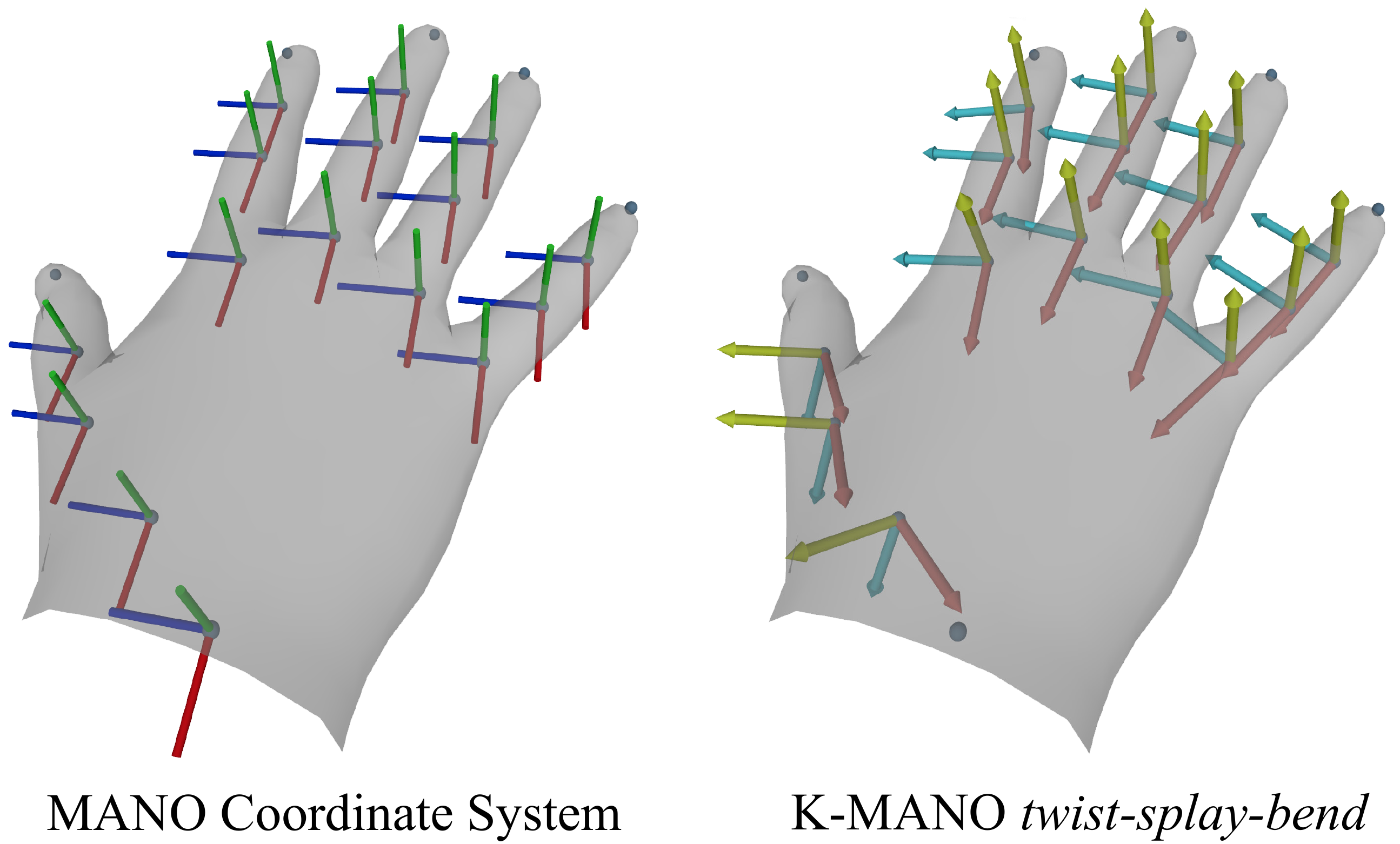}
   \caption{Visual comparison of MANO' s coordinate system to the proposed \textit{twist-splay-bend} system.}
   \label{fig:axis_adaptation}
\end{figure}

\begin{figure}
   \centering
   \includegraphics[width=1.0\linewidth]{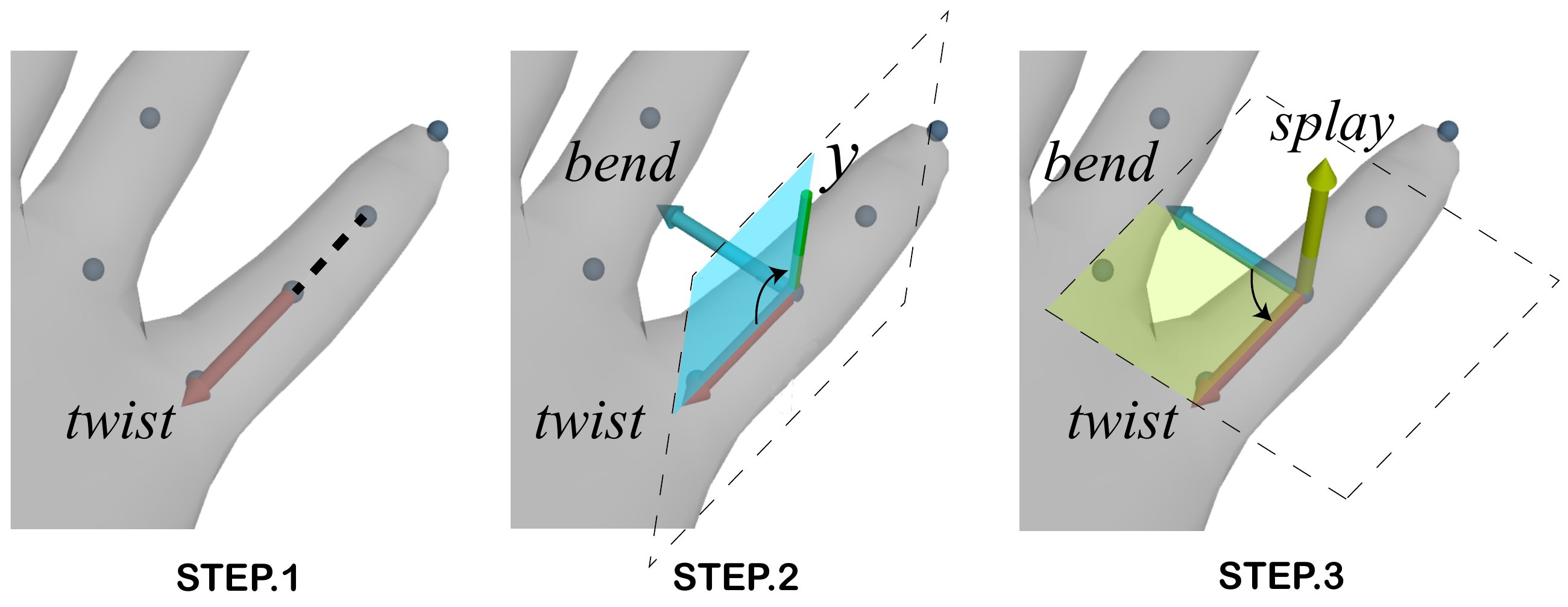}
   \caption{Illustration of converting MANO' s coordinates system to the proposed \textit{twist-splay-bend} system.}
   \label{fig:axis_generation}
\end{figure}

In this section, we introduce the proposed \textit{twist-splay-bend} frame of A-MANO. Both the original MANO \cite{romero2017embodied} and our A-MANO hand model are driven by the relative rotation at each articulation. To mitigate the pose abnormality, we apply constraints on the rotation \textit{axis-angle}\footnote{ Rotation cay be represented as rotating along an  \textit{axis} by an \textit{angle}.}.
We intend to decompose the rotation \textit{axis} into three components to the three axes of a Euclidean coordinate frame,
in which each component depicts the proportion of rotation along that axis.
Obviously, there have infinity choices of the three orthogonal axes.
MANO adopts 16 identical coordinate frames whose 3 orthogonal axes are not coaxial to the direction of the hand kinematic tree (Fig. \ref{fig:axis_adaptation} left).
Different from MANO, we follow the Universal Robot Description Format (URDF) \cite{lynch2017modern} that describe each articulation along the hand kinematic tree as a revolute joint\footnote{\url{https://en.wikipedia.org/wiki/Revolute_joint}}.
Nevertheless, a revolute joint only has one degree of freedom, which is not enough to drive the motion of a real hand.
Thus, we assign each articulation with three revolute joints, named as \textit{twist}, \textit{splay} and \textit{bend} (Fig. \ref{fig:axis_adaptation} right),

Here, we elaborate the conversion from the MANO's all identical coordinate system of to our \textit{twist-splay-bend} frame in three steps.
For each articulation, we first compute the \textit{twist} axis as the vector from the child of the  current joint to itself.
Then we employ MANO's $y$ (up) axis and derive the \textit{bend} axis that is calculated from cross product on the \textit{twist} and  $y$ axes.
Finally, we obtain the \textit{splay} axis by applying cross product on the \textit{bend} and \textit{twist} axes.
We illustrate the above procedures in Fig. \ref{fig:axis_generation}.

\subsection{Hand Subregion Assignment} \label{sec:subregion}

As introduced in main text \S \ref{sec:k_mano} (Anchors.), we divide the hand palm into 17 subregions, and interpolate the vertices in each subregion into representative \textit{anchor} /\ \textit{anchors}. In this part, we will firstly discuss how we assign hand vertices to several subregion.

\begin{figure}[t]
    \centering
    \includegraphics[width=0.99\linewidth]{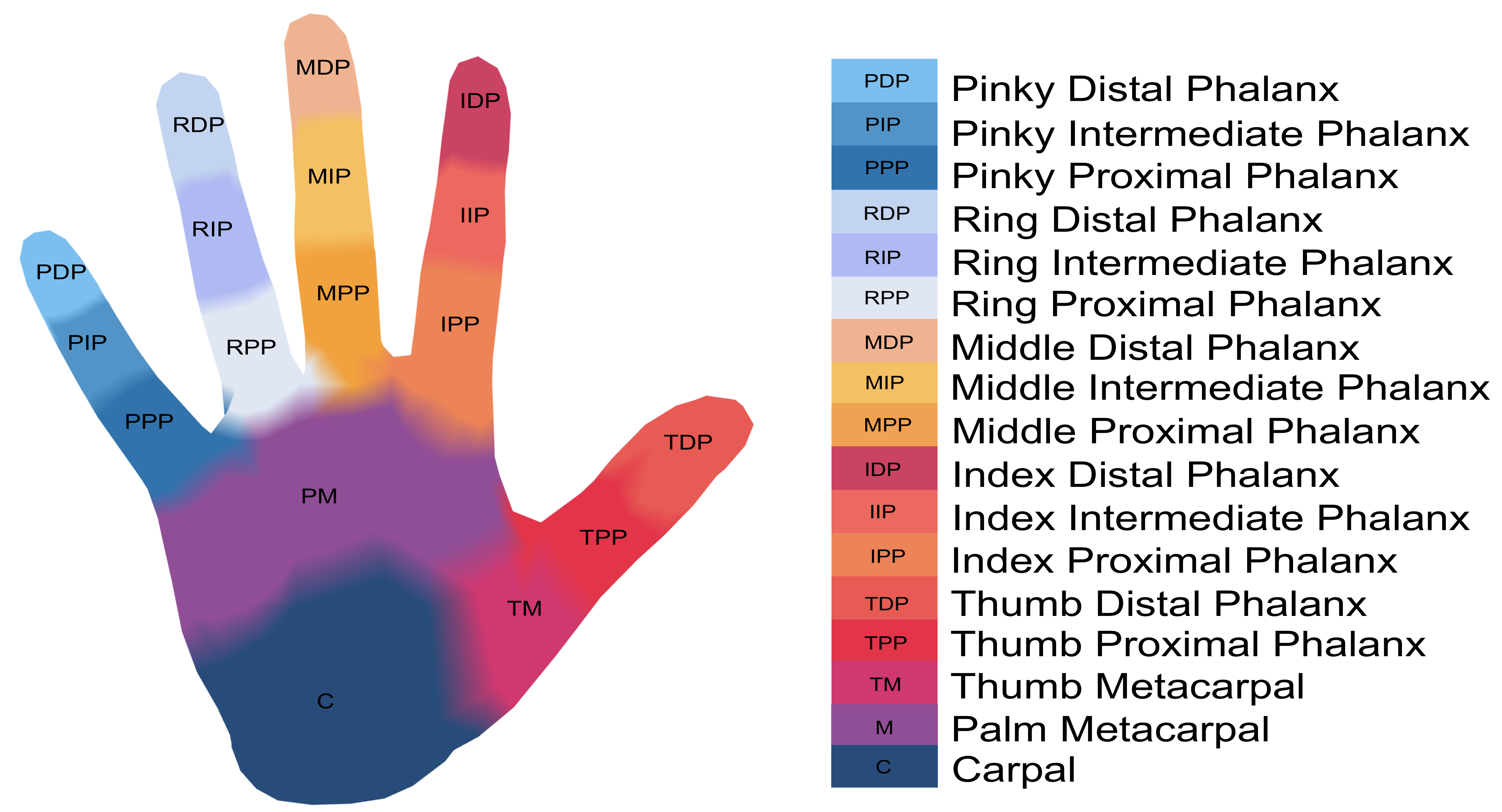}
    \caption{Hand subregions with names and abbreviations.}
    \label{fig:region_assignment}
\end{figure}

\begin{figure}[t]
    \centering
    \includegraphics[width=0.99\linewidth]{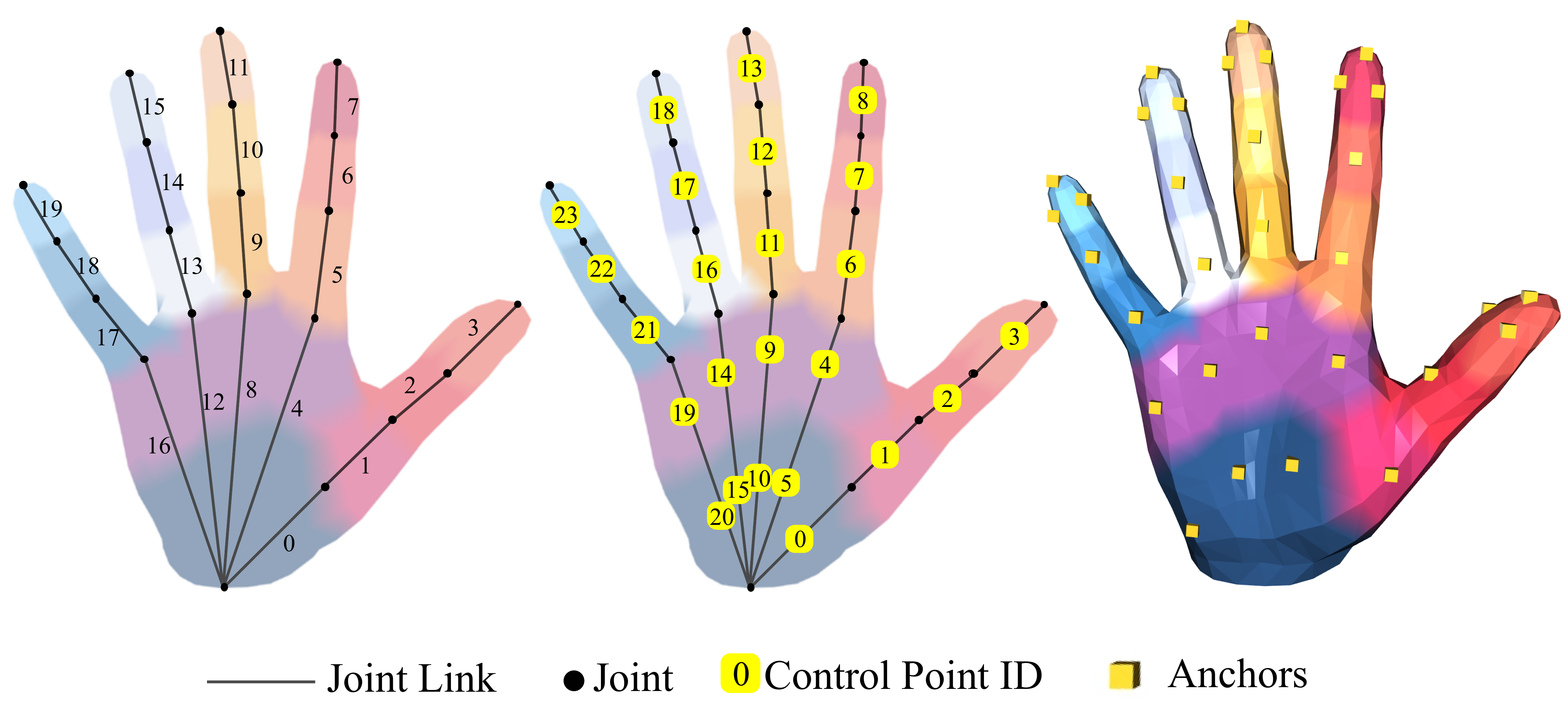}
    \caption{Left: joint links with ID; Middle: control points with ID; Right: anchors}
    \label{fig:anchor_coorespond}
\end{figure}

According to hand anatomy, the linkage bones consists of carpal bones, metacarpal bones, and phalanges, where phalanges can be further divided into three kinds: proximal phalanges, intermediate phalanges, and distal phalanges.
Here we assume the link between MANO joints are a counterpart of linkage bones on hand.
We now assign the vertices of MANO into 17 subregions based on the linkage bones.
The subregions' names and abbreviations are defined in Fig. \ref{fig:region_assignment}.
For clarity, we number the MANO links from 1 to 20 as illustrated in Fig. \ref{fig:anchor_coorespond} (left).

To assign the MANO vertices to its corresponding region, we need firstly assign the vertices to the link that lies inside the region.
This is achieved by \textit{control points}. For link 0-3, 5-7, 9-11, 13-15, 17-20, we set one control point at the midpoint of the link's ends, while for link 4, 8, 12, and 16, we set two control points at the upper and lower third of the link's ends. For clarity, we also number the control points from 0 to 23 as shown in Fig.\ref{fig:anchor_coorespond} (middle).
After a list of control points are obtained, we label each hand vertex to one of these control points by querying which control point it has the least distance from.
Finally, we merge the vertices that belong to control points 0, 5, 10, 15, and 20 to derive subregion of \textbf{\color[RGB]{144,78,150}Palm Metacarpal}, and merge those vertices that belong to control points 4, 9, 14, 19 to derive subregion of \textbf{\color[RGB]{40,76,121}Carpal} .

\subsection{Hand Anchor Selection}\label{sec:ancohor_selection}
Here we elaborate on how we select the \textit{anchors} based on the subregions and their control points.
To ensure these anchors can be used in a common optimization framework and keep their representative power during the process of optimization, we propose the following three protocols:
\begin{enumerate*}[label={\alph*)},font={\bfseries}]
    \item Anchors should be located on the surface of the hand mesh.
    \item Anchors should distribute uniformly on the surface of the region it represents.
    \item Anchors can be derived from hand vertices in a differentiable way.
\end{enumerate*}

Anchors are located on the surface of hand mesh (protocol \textbf{a}), so they must be located on some certain faces of the hand mesh.
We can use the vertices of the face on which hand anchors reside to interpolate the anchors' position.
Suppose the hand mesh has the form of $\mathbf{M}=(\mathbf{V},\mathbf{F})$,
where $\mathbf{V}$ is a set of all vertices and $\mathbf{F}$ is a set of all faces.
Considering one face $\bm{f} \in \mathbf{F}$ of mesh whose vertices are stored in order: $\bm{f} = \{i_k\}, \bm{v}_k = \mathbf{V}[i_k], k\in\{1,2,3\}$.
We can get two edges of that face: $\bm{e}_1 = \bm{v}_2 - \bm{v}_1, \bm{e}_2 = \bm{v}_3 - \bm{v}_1$.
Then the local position of the anchor $\tilde{\bm{a}}$ inside the face can be represented by linear interpolation of $\bm{e}_1$ and $\bm{e}_2$: $\tilde{\bm{a}} = x_1 \bm{e}_1 + x_2 \bm{e}_2$, where the $x_1, x_2$ are some weights.
Finally, the global position of the anchor $\bm{a}$ will be $\bm{a} = \bm{v}_1 + \tilde{\bm{a}} = \bm{v}_1 + x_1 \bm{e}_1 + x_2 \bm{e}_2 = (1 - x_1 - x_2) \bm{v}_1 + x_1 \bm{v}_2 + x_2 \bm{v}_3$.
During the optimization process, we can use the precomputed face $\bm{f}$ and  weights $x_1, x_2$, along with the predicted hand vertices $\mathbf{V}$ to calculate the position of all the anchors.
As the anchor is a linear combination of hand vertices, any loss that is applied to the anchors' position can be backpropagated to the vertices on the MANO surface, making the anchor-bases hand mesh differentiable.

We utilize control points introduced in \S\ref{sec:subregion} to derive anchors.
Since the anchor selection is independent of hand's configuration, we adopt a flat hand in the canonical coordinate system.
As illustrated in Fig.\ref{fig:anchor_coorespond} (middle, right), the control points are roughly uniformly distributed in each subregion. Each control point will correspond to an anchor of that subregion. The \textbf{\color[RGB]{40,76,121}Carpal} is an only exception: we select only 3 over 5 (ID: 5, 10, 20)  of the control points in the subregion of \textbf{\color[RGB]{40,76,121}Carpal} for anchor derivation.

To derive an anchor from a control point, we need to get one face (consist of 3 integers) and two weights.
\begin{enumerate*}[label={\arabic*)},font={\bfseries}]
   \item \textbf{\textit{Non-tip regions}. } For non-tip regions, we cast a ray that is originated from each control point in a certain subregion, and pointing to the palm surface.
   We retrieve the first intersection of the ray with hand mesh. This intersection will be the anchor that correspond to the control point, also the subregion.
   \item \textbf{\textit{Tip regions}. } For tip regions, we would select three anchors of each control point to increase the density of anchors in that subregion, as tip involves more contact information during manipulation. For the control point in tip subregions, we first cast a ray originated from the control point and get the intersection point on the hand mesh. Then a cone is created with the control point as apex, the intersection point as the base center, and a base radius.
   The base radius is estimated by the maximum distance of vertices in the subregion to their control point. Three generatrices equally distributed on cone surface are selected as new ray casting directions. We cast three rays from the control point in the direction of the three generatrices and retrieve the intersection points with hand mesh. These intersection points will be selected as anchors to that control point in the fingertip regions.
\end{enumerate*}

\section{Spring's Elasticity}\label{suppsec:springs}
\subsection{Elastic Energy Analysis}\label{sec:elast_energy_analysis}

\begin{figure}[!t]
    \centering
    \includegraphics[width=\linewidth]{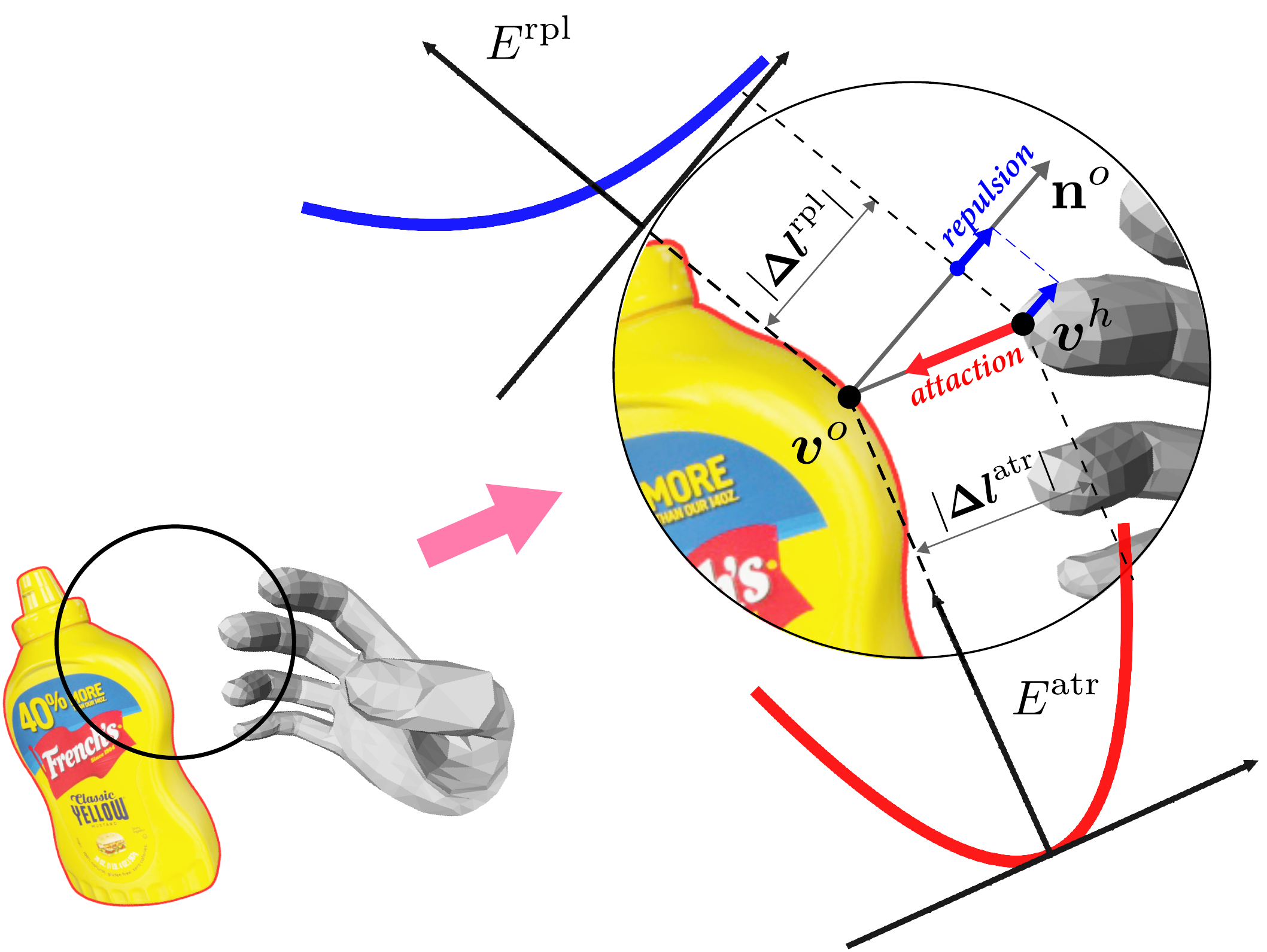}
    \caption{Illustration of the elastic energy \wrt a pair of hand-object vertices. }
    \label{fig:attract-repulsion_balance}
\end{figure}

Here we illustrate elastic energy between a pair of points $\bm{v}_i^h$ and $\bm{v}_j^o$, denoting one vertex on hand surface and another vertex on object surface respectively. The vertex on object surface binds with a vector $\mathbf{n}_j^o$ representing the normal direction at this vertex (also the direction of repulsion).
Then we compute the offset vector $\bm{\Delta l}_{ij}^{\rm{atr}} = \bm{v}_i^h - \bm{v}_j^o$, and the projection of the offset vector on object normal $\mathbf{n}_j^o$: $| \bm{\Delta l}_{ij}^{\rm{rpl}} | = (\bm{v}_i^h - \bm{v}_j^o) \cdot \mathbf{n}_j^o$. $| \bm{\Delta l}_{ij}^{\rm{rpl}} |$ is positive if $\bm{v}_i^h$ falls outside the object, and negative if $\bm{v}_i^h$ falls inside the object.
We use an exponential function here to provide magnitude and gradient heuristic for optimizer:
\begin{enumerate*}[label={\alph*)},font={\bfseries}]
\item the less $|\bm{\Delta l}_{ij}^{\rm{rpl}}|$ is, the more $\bm{v}_i^h$ penetrates into the object. The gradient of repulsive energy will be an exponential increasing function of $\bm{\Delta l}_{ij}^{\rm{rpl}}$.
\item when $\bm{v}_i^h$ intersects into the object, both the repulsion and the attraction will push $\bm{v}_i^h$ towards the surface; when $\bm{v}_i^h$ is outside the object, the attraction and repulsion will point to opposite directions, leading to a balance point outside but in the vicinity to the object's surface.
We provide an intuitive illustration in Fig. \ref{fig:attract-repulsion_balance}.
\end{enumerate*}

\subsection{Anchor Elasticity Assignment}\label{sec:anchor_assignment}

As discussed in main text \S \ref{sec:cpf} (Annotation of the Attractive Springs), we treat the elasticity of the \textit{attractive} spring as the network prediction.
Here, we shall provide the annotation heuristics of the \textit{attractive} spring $\hat{k}^{\rm{atr}}$
First, we set the anchor $\bm{a}_i$ - vertex $\bm{v}^o_j$ pair with ground-truth distance $| \bm{\Delta} \hat{\bm{l}}^{\rm{atr}}_{ij} | > 20 mm$ as invalid contact and has $\hat{k}^{\rm{atr}}_{ij}=0$.
Second, for those anchor-vertex pairs within the distance threshold 20 $mm$, an inverse-proportional $\hat{k}^{\rm{atr}}_{ij}$ is assigned according to the $|\bm{\Delta} \hat{\bm{l}} ^{\rm{atr}}_{ij}|$:
\begin{equation}
   \hat{k}^{\rm{atr}}_{ij} = 0.5 * \cos \big (\frac{\pi}{s} * |\bm{\Delta} \hat{\bm{l}}^{\rm{atr}}_{ij} | \big ) + 0.5
\end{equation}
where the scale factor $s = 20 \ mm$.

To note, we do not have a strict requirement on the function of $\hat{k}^{\rm{atr}}_{ij}$. Any other functions should also work when satisfying:
\begin{enumerate*}[label={\alph*)},font={\bfseries}]
\item $k = 1$ when $|\bm{\Delta l}|=0$;
\item $k$ is inverse proportional to $|\bm{\Delta l}|$ in the range of 0 to 20 $mm$;
\item $k$ is bounded by 0 and 1. The choice of cosine function is simply due to its smoothness.
\end{enumerate*}

\section{HO3D Dataset}\label{suppsec:ho3d}
\subsection{Analysis and Selection}\label{sec:ho3d_selection}
\begin{figure}[t]
   \centering
   \includegraphics[width=1.0\linewidth]{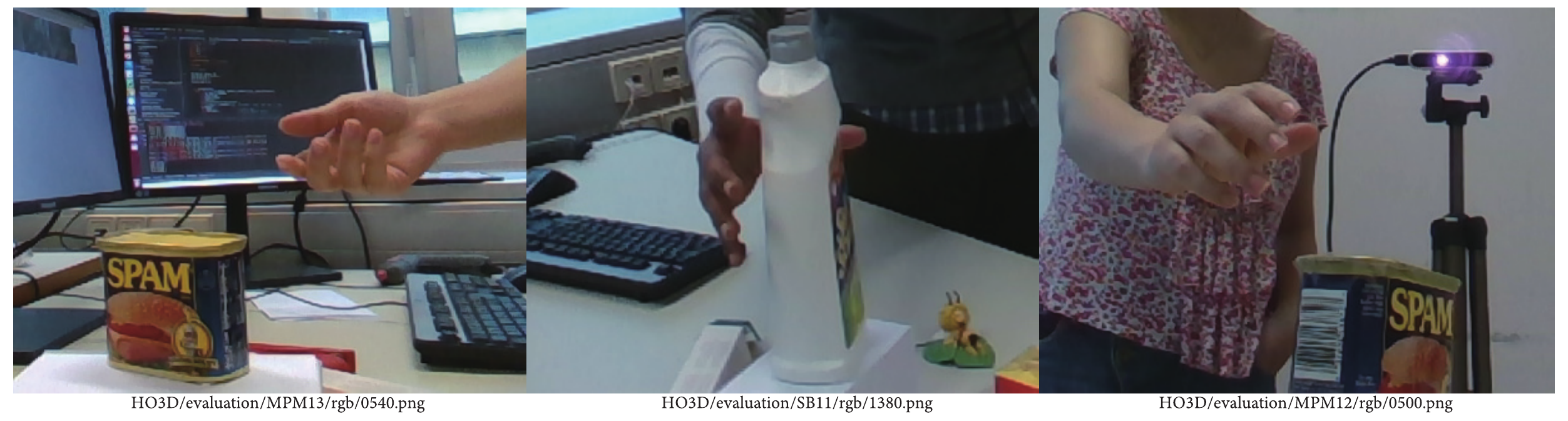}
   \caption{Unsuitable samples in HO3Dv2 testing set.}
   \label{fig:unsuitable_samples}
\end{figure}

As we mentioned in the main text \S \ref{sec:dataset}, several samples in the HO3D testing set do not suit for evaluating MIHO. Firstly, since GeO requires the predicted 6D pose of the known objects, all the grasps of the \textit{pitcher} have to be removed. Secondly, many interactions of hand and objects in the testing set are not stable. For example, sliding the palm over the surface of a \textit{bleach cleanser bottle}, may cause a strange contact and mislead the optimization in GeO. Therefore, we only select the grasps that can pick up the objects firmly. We show several unsuitable samples in Fig. \ref{fig:unsuitable_samples}. Table.\ref{table:ho3dv2} shows our final selection on HO3Dv2 test set, as we called HO3Dv2$^-$.

\begin{table}[h]
    \begin{small}
    \centering
        \begin{tabular}[width=\columnwidth]{c|c}
            \toprule
        Sequences   &  Frame ID \\
            \midrule
        SM1  &  All \\
        MPM10-14  &  30-450, 585-685 \\
        SB11 & 340-1355, 1415-1686 \\
        SB13 & 340-1355, 1415-1686 \\
            \bottomrule
        \end{tabular}
    \vspace{10 pt}\caption{\textbf{HO3Dv2$^-$ selection.} We select 6076 samples in the HO3Dv2 test set to evaluate MIHO.}
    \label{table:ho3dv2}
    \end{small}
\end{table}

\begin{figure*}[t]
    \centering
    \includegraphics[width=1.0\linewidth]{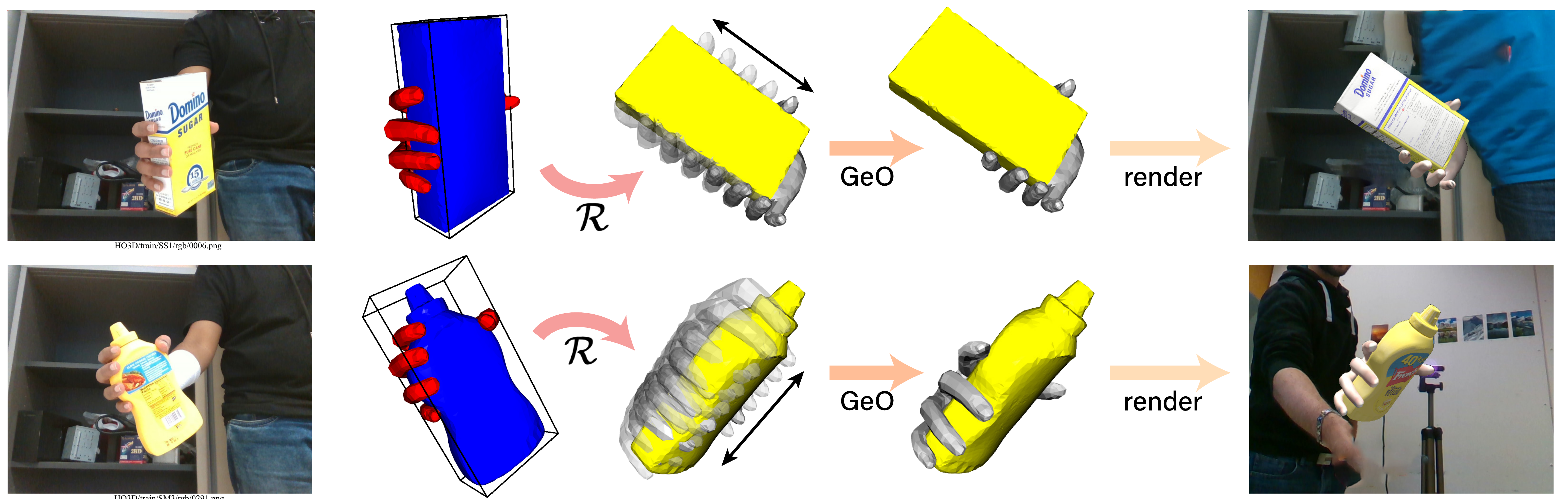}
    \caption{\textbf{HO3D \cite{hampali2019ho} Dataset augmentation.} We demonstrate the process of generating synthetic training images. $\mathcal{R}$ stands for the random transformation. }
    \label{fig:data_augmentation}
\end{figure*}

\subsection{Data Augmentation}\label{sec:ho3d_data_augmentation}
We augment the training sample in HO3Dv1 in terms of poses and grasps.
\begin{enumerate*}[label={\alph*)},font={\bfseries}]
\item To generate more poses, we firstly randomize a disturbance transformation to the hand and object poses in the object canonical coordinate system. Then, we apply the disturbance on the hand and object meshes and render these meshes to image by a given camera intrinsic.
\item To generate more grasps, we fit more stable grasps around the object. Specifically, as we show in Fig. \ref{fig:data_augmentation}, the generation procedure is achieved by 2 steps: 1) Manually move the hand around the tightest bounding cuboid of the object. 2) Refine the hand pose in the proposed GeO. Since the \textit{attractive} springs in CPF are unavailable here, we replace the attraction energy in main text Eq. 3  with the $\mathcal{L}_A$ in \cite{hasson2019learning} Eq. 4, and retain the repulsion energy and the anatomical cost. The optimization process of grasping generation can be expressed as:

\end{enumerate*}
\begin{equation}
  {\mathcal{\hat{V}}}^{h} \longleftarrow  \mathop{argmin}_{(\bm{P}_w , \bm{R}_j)} (\mathcal{L}_{A} + E^{\rm{rpl.}} + \mathcal{L}_{\rm{anat}})
  \label{aug_GeO}
\end{equation}

\section{Experiments and Results}\label{suppsec:expm}
\subsection{Implementation Details}\label{sec:more_impl_details}

In this section we provide more implementation details about the HoNet, PiCR, and GeO module.

\paragraph{HoNet.} The HoNet module employs ResNet-18 \cite{he2016deep} backbone initialized with ImageNet \cite{deng2009imagenet} pretrained weights.
For FHB and HO3Dv2 dataset, we use the pretrained weights released from \cite{hasson2020leveraging}. For the HO3Dv1 dataset, we train the HoNet with Adam solver and a constant learning rate of $5\times 10^{-4}$ in total 200 epochs.

\paragraph{PiCR.}
The PiCR module employs a Stacked Hourglass Networks \cite{newell2016stacked} (with 2 stacks) as backbone, a PointNet \cite{qi2017pointnet} as the point encoder, and three multi-layer perceptrons as heads.
The image features yield from the two hourglass stacks are gathered together and sequentially fed into the PointNet encoder and three heads. While the loss is computed over the sum of two rounds prediction, both PointNet encoder and the three heads have only one instance throughout PiCR module. At the evaluation stage, we only use the image features from the last hourglass stack to get the prediction from three heads.

We train the PiCR module with two stages.
\begin{enumerate*}[label={\arabic*)},font={\bfseries}]
   \item \textbf{\textit{Pretraining.}} We pretrain the PiCR module with the input image and the ground-truth object mesh in camera space. The ground-truth object mesh are disturbed by a minor rotation and translation shift. We employ Adam solver with an initial learning rate of $1 \times 10^{-3}$, decaying 50\% every 100 epochs. The total epochs during pretraining stage is 200.
   \item \textbf{\textit{Fine-tuning.}} At the fine-tuning stage, we feed PiCR module with the object vertices predicted from HoNet. The HoNet's weights is freezed during PiCR fine-tuning. We employ Adam solver and set the initial learning rate in fine-tuning stage as $5\times 10^{-4}$, decayed to 50\% every 100 epochs, and finished at 200 epochs.
\end{enumerate*}
In both stages, we set the training mini-batch size to 8 per GPU, and a total of 4 GPUs are used.

\paragraph{GeO.}
The GeO is a fitting module based on the non-linear optimization. For each sample, we minimize the cost function in 400 iterations, with a initial learning rate of $1\times 10^{-2}$, reduced on plateau that the cost function has stopped decaying in 20 consecutive iterations.
We implement GeO in PyTorch thanks for its auto derivative, and an Adam solver is employed when updating the arguments. To note, GeO can also support any other optimization toolbox.

\subsection{Ablation Study}\label{sec:more_ablation_study}
As referred in main text \S \ref{sec:ablation} (Ablation Study), this section contains another three ablation studies. all the following experiments are under the \textit{\textbf{hand-object}} setting.

\paragraph{The Impact of the $k^{\rm{rpl}}$. }
While the elasticity $k^{\rm{atr}}$ of the \textit{attractive} springs are predicted in PiCR, the elasticity
$k^{\rm{rpl}}$ of those \textit{repulsive} strings are empirically set to $1 \times 10 ^{-3}$.
In order to measure the impact of the magnitude of  $k^{\rm{rpl}}$ on repulsion, we test our GeO with seven experiment settings in which the $k^{\rm{rpl}}$ is set to $\{ 0.2,\ 0.6,\ 1.0,\ 1.4,\ 2.0,\ 4.0,\ 8.0\} \times 10^{-3}$, respectively.  The experiment with $k^{\rm{rpl}} = 1 \times 10^{-3}$ is in accord with the default experiment in main text. As shown in Tab. \ref{table:ablation_rpl}, while the large $k^{\rm{rpl}}$ can  reduce the solid interpenetration volume, it may also push the attraction apart thus is not preferable in the reconstruction metrics: hand MPVPE and object MPVPE.

\begin{table}[h]
   \rowcolors{4}{Gray}{}
   \begin{footnotesize}
   \begin{center}
       \begin{tabular}{c|ccccc}
       \toprule
       \multirow{2}{*}{$k^{\rm{rpl}}$} & \multicolumn{5}{c}{{Scores}} \\
       \cmidrule(r){2-6}
        & HE $\downarrow$ & OE $\downarrow$ & PD  $\downarrow$ & SIV $\downarrow$ & DD $\downarrow$ \\
       \midrule
       $2.0\times10^{-4}$ & 19.49 & 21.57 & 17.77 & 13.22 & 20.85 \\
       $6.0\times10^{-4}$ & 19.51 & 21.57 & 17.22 & 12.40 & 21.63 \\
       $\mathbf{1.0\times10^{-3}}$ & \textbf{19.54} & \textbf{21.57} & \textbf{16.92} & \textbf{11.76} & \textbf{22.41} \\
       $1.4\times10^{-3}$ & 19.59 & 21.58 & 16.75 & 11.00 & 23.24 \\
       $2.0\times10^{-3}$ & 19.69 & 21.59 & 16.41 & 10.09 & 24.55 \\
       $4.0\times10^{-3}$ & 20.03 & 21.63 & 15.09 & 7.65 & 29.33 \\
       $8.0\times10^{-3}$ & 20.95 & 21.92 & 12.86 & 4.27 & 40.79 \\
       \bottomrule
       \end{tabular}
   \end{center}
   \end{footnotesize}
   \caption{\textbf{Ablation results:} the impact of the magnitude of $k^{\rm{atr}}$. HE stands for hand mean per vertex position error ($mm$); OE stands for object mean per vertex position error ($mm$); PD stands for penetration depth ($mm$); SIV stands for solid intersection volume ($cm^3$); D stands for disjointedness distance ($mm$).}
   \label{table:ablation_rpl}
\end{table}

\paragraph{A-MANO with PCA Pose.}
Since the MANO can also be driven by the PCA components of joint rotation, we further conduct experiments to demonstrate the superiority of our full MANO ( MANO with 15 relative joint rotations) over the PCA MANO (MANO with 15 PCA components of rotations). Tab. \ref{table:ablation_pca} shows that our full MANO can achieve a notable decrease in the hand MPVPE. We attribute this to the fact that the PCA MANO tends to recovery a hand that is inclined to the mean flat pose, while our full version imposes higher flexibility on the hand pose.

However, fitting on the 15 rotations in forms of $\mathfrak{so}(3)$ brings $15 \times 3 = 45$ degree of freedoms, which is less stable against pose abnormality. Hence in order to fully exploit the advantages when fitting on the rotations of 15 joints, we have to combine the anatomical constrains with it.

\begin{table}[h]
    \footnotesize
    \renewcommand{\arraystretch}{1}
    \begin{center}
        \begin{tabular}{c|ccccc}
        \toprule
        \multirow{2}{*}{Settings} & \multicolumn{5}{c}{{Scores}} \\
        \cmidrule(r){2-6}
         & HE $\downarrow$ & OE  $\downarrow$ & PD  $\downarrow$ & SIV  $\downarrow$ & DD  $\downarrow$ \\
        \midrule
        \textbf{Full MANO} & \textbf{19.54} & \textbf{21.57} & \textbf{16.92} & \textbf{11.76} & \textbf{22.41} \\
        PCA MANO & 23.32 & 24.41 & 22.47 & 11.90 & 26.72 \\
        \bottomrule
        \end{tabular}
    \end{center}
    \caption{\textbf{Ablation results:} the MANO with PCA pose.}
    \label{table:ablation_pca}
\end{table}

\begin{figure}[t]
   \centering
   \includegraphics[width=0.9\linewidth]{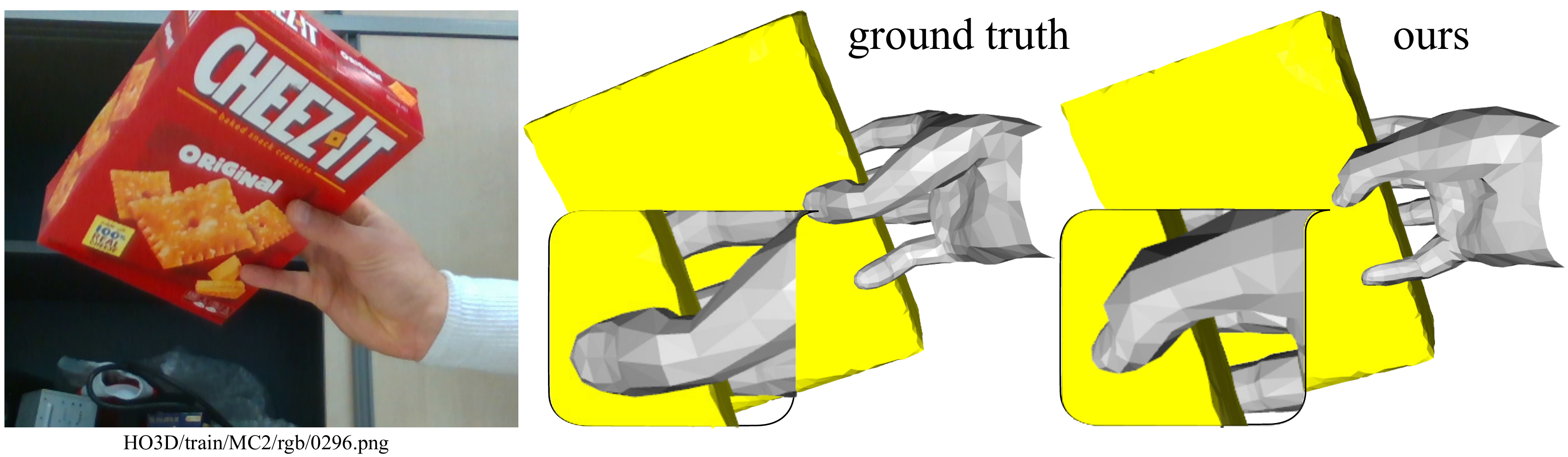}
   \caption{Example to show that our A-MANO can mitigate the unwanted twist (see thumb) exhibited in ground-truth. }
   \label{fig:twist_correction}
\end{figure}

\paragraph{Unwanted Twist Correction.}
In this part, we show the effectiveness when fitting the 15 rotations with anatomical constrains. We observe an unwanted twist of thumb in the ground-truth pose of HO3Dv1 testing set. As shown in Fig. \ref{fig:twist_correction}, since A-MANO imposes constraints on the \textit{twist} component of the rotation axis, it can achieve a more visually pleasing result in such case.

\section{More Qualitative Results}\label{suppsec:morequal}
\begin{figure*}[t]
   \centering
   \includegraphics[width=0.93\linewidth]{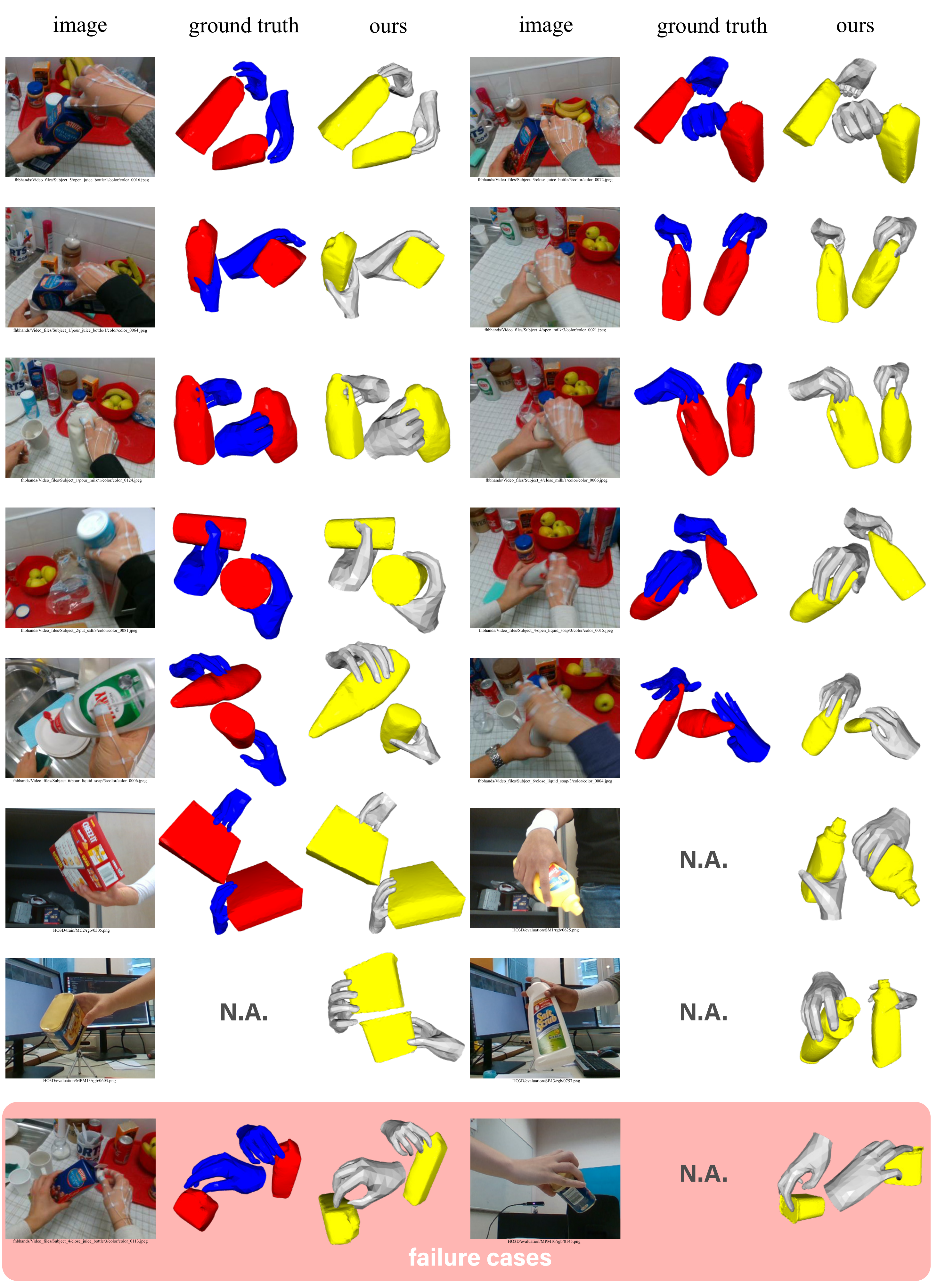}
   \caption{\textbf{Qualitative results on FHB \cite{garcia2018first}, HO3Dv1\cite{hampali2019ho} and HO3Dv2$^{-}$ \cite{hampali2020honnotate} datasets.} The last row shows the failure cases.}
   \label{fig:full_qualitative}
\end{figure*}

We demonstrate the qualitative results of MIHO in Fig. \ref{fig:full_qualitative} on both the FHB \cite{garcia2018first} and HO3D dataset \cite{hampali2020honnotate}. Note that the ground truth of the test set in HO3Dv2$^{-}$ \cite{hampali2020honnotate} is not available.

\end{appendix}

\end{document}